\documentclass[pdflatex,sn-mathphys-num]{sn-jnl}
\usepackage[utf8]{inputenc}
\usepackage{xcolor}
\usepackage{listings}
\hypersetup{hypertexnames=false,bookmarksdepth=3,
  pdftitle={Graph Memory Transformer (GMT)},
  pdfauthor={Nicola Zanarini; Niccol\`o Ferrari; Evelina Lamma}}
\usepackage{url}
\newcommand{\burl}[1]{\url{#1}}
\newcommand{\arxivurl}[1]{\url{#1}}
\usepackage{changepage}
\usepackage{graphicx}
\graphicspath{{figures/}}
\usepackage{mathtools}
\usepackage{amsmath}
\usepackage{amssymb}
\usepackage{bm}
\usepackage{multirow}
\usepackage{manyfoot}
\DeclareNewFootnote{A}[alph]
\usepackage{multicol}
\usepackage{algorithm}
\usepackage{algpseudocode}
\usepackage{acronym}
\usepackage{booktabs}
\usepackage{array}
\usepackage{longtable}
\usepackage[stretch=10,shrink=10]{microtype}
\acrodef{GMT}{Graph Memory Transformer}
\acrodef{FFN}{Feed-Forward Network}
\acrodef{MLP}{Multi-Layer Perceptron}
\acrodef{EMA}{Exponential Moving Average}

\begin{document}

\title{Graph Memory Transformer (GMT)}

\author[1]{\fnm{Nicola} \sur{Zanarini}}\email{nicola.zanarini@hotmail.com}
\author*[2,3]{\fnm{Niccol\`o} \sur{Ferrari}}\email{niccolo.ferrari@unife.it}
\author[2]{\fnm{Evelina} \sur{Lamma}}

\affil[1]{\orgname{Bonfiglioli Engineering s.r.l.},
  \orgaddress{\city{Ferrara}, \country{Italy}}}
\affil[2]{\orgdiv{Department of Engineering}, \orgname{University of Ferrara},
  \orgaddress{\street{Via Saragat 1}, \city{Ferrara}, \postcode{44122}, \country{Italy}}}
\affil[3]{\orgname{NAIS s.r.l.},
  \orgaddress{\street{Via Maria Callas 4}, \city{Bologna}, \postcode{40131}, \country{Italy}}}

\AddToHook{cmd/printabstract/before}{%
\begingroup\renewcommand{\thefootnoteA}{}%
\footnotetextA{\textbf{Acknowledgments.}
This work was conducted as independent research using local computational resources.
The authors thank the open-source machine-learning community for the software
libraries, public datasets, and evaluation tools that made the experiments possible.}%
\endgroup}

\abstract{%
\paragraph*{Purpose}
We investigate whether an explicit learned memory graph can replace the \ac{FFN}
sublayer in a decoder-only transformer while preserving the surrounding autoregressive
architecture.
\paragraph*{Methods}
In the proposed \ac{GMT}, a memory cell routes token representations
over a learned bank of centroids connected by a learned directed transition matrix,
while causal self-attention remains unchanged. In the base v7 instantiation, each of 16
transformer blocks contains 128 centroids and a $128\times128$ edge matrix. Each cell
uses gravitational source routing and token-conditioned target selection to return a
gated displacement from the estimated source memory state toward the target memory
state, rather than a retrieved value.

\paragraph*{Results}
The resulting decoder-only language model has 82.2M trainable parameters and no dense
\ac{FFN} sublayers; the evaluated dense GPT-style baseline has 103.0M. It trains stably
and exposes centroid usage, transition structure, and source-to-target movement for
direct inspection in the forward computation. It trails the larger baseline in
validation loss and perplexity (3.5995/36.58 vs.\ 3.2903/26.85), while showing close
zero-shot benchmark behavior under the evaluated setting.
\paragraph*{Conclusion}
These results support the
viability and structural interpretability of replacing dense within-token transformation
with graph-mediated memory navigation, without claiming state-of-the-art performance.
Broader scaling, optimized kernels, and more extensive benchmark evaluation remain for
future work.
}


\maketitle

\noindent\textbf{Code:} \url{https://github.com/Nemesis533/GMT-GraphMemoryTransformer}

\section{Introduction}
\label{sec:intro}

\subsection{Motivation}

In transformer language models \cite{vaswani2017attention}, \acp{FFN} provide much of the model's transformation capacity, functioning as implicit key-value memories \cite{geva2021transformer}. Yet they remain difficult to inspect directly. Once trained, they apply rich nonlinear updates without exposing an explicit intermediate structure that can be examined, queried, or modified in a principled way. It is therefore often unclear which internal concepts are being used, how they are organized, or how a token is moved from one representation to the next inside the block \cite{meng2022locating,rai2024practical}.

\ac{GMT} explores whether part of this computation can instead be carried by an explicit internal memory \cite{weston2015memory,sukhbaatar2015endtoend,lample2019large,wu2022memorizing,berges2024memory,jaiswal2026memoryllm}. The aim is to replace the opaque \ac{FFN} sublayer with identifiable, structurally legible components, following work on interpretability and explainability \cite{olah2020zoom,cammarata2020thread,rai2024practical}. Exposing centroid usage, transition structure, and token-dependent movement through the memory graph for direct inspection makes this sublayer less opaque than a conventional dense \ac{FFN}. The goal is a more structured, analyzable internal memory, not full transparency of transformer computation \cite{yordanov2026prototype,jaiswal2026memoryllm}.

\subsection{Core Idea of GMT}

\ac{GMT} preserves the overall transformer structure, including causal self-attention's role in integrating contextual information across the sequence \cite{vaswani2017attention}. It changes only the branch normally occupied by the \ac{FFN}, whose computational role has been studied extensively \cite{geva2021transformer,meng2022locating}. A memory cell uses a learned bank of centroids connected by a learned transition graph to produce the block update in place of a dense channel-wise transformation \cite{weston2015memory,sukhbaatar2015endtoend,berges2024memory}.

The memory cell treats the current token representation as a point in a learned conceptual space. It first identifies a source region as a soft distribution over centroids rather than a hard slot, then combines learned graph transitions with token-dependent context to determine a target region better matched to the transformation required at that stage. Its output is not a retrieved value in the usual key-value sense, but a displacement: the difference between the estimated source memory state and the target memory state selected for the token. This displacement is fed back into the residual stream.

In a conventional \ac{FFN}, the transformation is encoded implicitly in dense weight matrices and is difficult to decompose into meaningful intermediate operations \cite{geva2021transformer}. \ac{GMT} expresses the same transformation as movement between persistent centroid landmarks, with graph edges encoding preferred transitions. The update remains continuous and differentiable, while its components are more directly inspectable than those of a standard dense \ac{MLP} \cite{olah2020zoom,rai2024practical,yordanov2026prototype}. Memory thus reinterprets the \ac{FFN}'s computation inside the block rather than augmenting a standard transformer with a separate memory component. The expressive role of the \ac{FFN} is preserved in a structured, analyzable organization, potentially more amenable to post hoc inspection or modification.

\subsection{Why This Could Matter}

Centroids and graph transitions may clarify the model's internal computation by providing concrete objects for studying which memory regions are active, how tokens move between them, whether transitions specialize, and how their organization changes during training \cite{olah2020zoom,cammarata2020thread,rai2024practical}. They also open a possible route to more direct intervention. Such interventions may remain difficult in practice, but identifiable memory components are conceptually more receptive to post hoc inspection and modification than computation diffused across dense matrices \cite{meng2022locating,zhang2024comprehensive}.

\subsection{Main Contribution}

This paper introduces \ac{GMT} v7, a decoder-only transformer that replaces each block's \ac{FFN} with a memory cell. The contribution is both architectural---source routing, graph-mediated target selection, and a gated displacement readout---and operational, specifying the auxiliary objectives, online write-back rule, and centroid-maintenance procedures that keep the learned memory structure usable during training. More broadly, the work asks whether part of dense within-token computation can be usefully recast as explicit, inspectable memory navigation under the same modeling constraints, rather than relying on unconstrained channel mixing alone.

\subsection{Paper Roadmap}

Section~\ref{sec:related} reviews transformer internals, memory-augmented models, and interpretability-oriented alternatives. Section~\ref{sec:method} describes the architecture, routing, displacement readout, training objectives, and memory maintenance. Experimental setup and results follow, then a discussion of the architecture's viability, limitations, and broader significance.

\section{Related Work}
\label{sec:related}

\subsection{Transformer Feed-Forward Computation}

The transformer \ac{FFN} is often described operationally as a position-wise nonlinear
transformation, but several analyses suggest that its role is richer than simple channel
mixing. Geva et al. \cite{geva2021transformer} interpret transformer feed-forward layers
as key-value memories, in which the first projection detects patterns and the second
projection writes associated information back into the residual stream. Knowledge editing
methods similarly provide evidence that factual associations can be localized and modified
within feed-forward parameters \cite{meng2022locating,zhang2024comprehensive}. These
observations motivate \ac{GMT}'s replacement of the sublayer with a memory block that
makes the \ac{FFN}'s implicit memory function explicit and more structurally interpretable through
directly inspectable centroids and directed transitions.

\subsection{Memory-Augmented Neural Architectures}

Early memory-augmented models---Memory Networks
\cite{weston2015memory}, End-to-End Memory Networks \cite{sukhbaatar2015endtoend}, Neural
Turing Machines \cite{graves2014neural}, and Differentiable Neural Computers
\cite{graves2016hybrid}---introduced mechanisms for reading from and writing to explicit
memory stores. A related thread reframes attention itself as continuous associative
retrieval over learned patterns, culminating in modern Hopfield networks
\cite{ramsauer2021hopfield}. More recent language-model architectures incorporate large
external or semi-external memories through product-key layers, nearest-neighbor retrieval,
memorizing transformers, and memory layers at scale
\cite{lample2019large,khandelwal2020generalization,wu2022memorizing,berges2024memory}.

In \ac{GMT}, memory is part of the layer's internal computation rather than an auxiliary
retrieval component appended to a conventional transformer. Source-centroid routing,
propagation through the learned transition graph, and the resulting displacement update
replace the block-level \ac{FFN} transformation.

\subsection{FFN Replacement and Prototype Variants}

Work on alternatives to the dense, opaque \ac{FFN} includes a long-standing approach
that replaces or augments it through routing: Mixture-of-Experts models
\cite{jacobs1991adaptive,shazeer2017outrageously} and
Switch Transformers \cite{fedus2022switch} gate token states to specialist \ac{FFN}
sub-modules. \ac{GMT} likewise reorganizes the \ac{FFN} around routing, but routes to
passive centroids connected by a learned transition graph rather than a bank of
specialist \ac{MLP}s.

Like \ac{GMT}, MemoryLLM \cite{jaiswal2026memoryllm} treats the \ac{FFN} as more than
a generic nonlinear projection. It stays closer to the original form, decoupling the
\ac{FFN} from self-attention so that its token-wise retrieval behavior can be studied and
precomputed in isolation. \ac{GMT} instead replaces the residual branch with an active
memory cell whose routing, transition, and displacement variables are part of the forward
computation.

MemoryFormer \cite{ding2024memoryformer} removes most fully connected computations and
replaces them with memory lookup tables of discrete vectors. Its goal is primarily
computational: retrieved vectors are combined to
approximate the effect of linear projection at lower cost. In \ac{GMT}, memory instead
has a representational role: the edge matrix defines directed transitions between
layer-local centroid landmarks, and the update is a displacement between source and target
memory states, not a lookup approximation to matrix multiplication.

Shehzad et al. \cite{shehzad2026graphtransformers} survey graph transformers for node-,
edge-, and graph-level tasks, covering graph inductive biases, attention mechanisms,
scalability strategies, and pretraining. Their relationship to \ac{GMT} is conceptual,
not architectural: both combine transformers with graph structure, but at different
levels of representation. \ac{GMT}'s input remains an autoregressive token sequence;
its graph is a learned internal memory topology within the \ac{FFN} replacement branch.

Prototype Transformer \cite{yordanov2026prototype} is the closest conceptual neighbor
to \ac{GMT}'s centroids. It builds an autoregressive language model around learned
prototype vectors that can acquire concept-like roles during training. Both architectures
favor legible internal objects, but Prototype Transformer makes them central communication
objects for the whole model, whereas \ac{GMT} retains ordinary causal attention and
places centroids locally in each block's second residual branch. This narrower intervention
asks whether the transformation normally assigned to the \ac{FFN} can be expressed as
graph-structured movement through memory.

\subsection{Interpretable-by-Design Alternatives}

Work on mechanistic interpretability and models designed to expose more structured
internal representations \cite{olah2020zoom,cammarata2020thread,rai2024practical} motivates a
design principle: stable internal objects, identified by their architectural role and
reused across inputs, allow their usage and dynamics to be studied more directly than
activations in an unconstrained dense sublayer. This does not make transformer computation
fully transparent.

\ac{GMT} gives these objects relational structure: centroids provide persistent memory
locations, and the edge matrix represents directed transition tendencies between them.
Its explainability claim is structural, not absolute. \ac{GMT} remains a neural language
model, but its \ac{FFN} replacement exposes analyzable intermediate quantities: source
routing, target selection, and displacement.

\section{Method}
\label{sec:method}

\subsection{Overall Architecture}

\ac{GMT} is a decoder-only transformer language model \cite{brown2020language} trained for next-token prediction. It retains the standard tokenizer interface and sums token and positional embeddings before passing them through a stack of causal transformer blocks. The final hidden states are normalized and projected to next-token logits by a language-model head whose weights are tied to the input embedding matrix \cite{press2017using}.

Each block uses a pre-normalization residual design \cite{xiong2020layer}, with causal self-attention \cite{vaswani2017attention} followed by a memory cell in place of the standard \ac{FFN}. The memory cell estimates a source distribution over layer-specific centroids, propagates it through a learned directed transition matrix, and combines the propagated distribution with token-conditioned target scores to form a target distribution. A gated displacement from the source memory state to the target memory state is then added to the residual stream.

\subsection{Transformer Block with Memory Cells}

Let $\mathbf{x}^{(\ell)} \in \mathbb{R}^{T \times H}$ denote the sequence of hidden states entering layer $\ell$, where $T$ is the sequence length and $H$ is the hidden dimension. We omit the batch dimension. The block preserves the transformer's residual stability and causal organization. Its attention branch applies causal self-attention \cite{vaswani2017attention} in the standard pre-normalized residual form \cite{xiong2020layer}:
\begin{equation}
\mathbf{h}^{(\ell)} = \mathbf{x}^{(\ell)} +
\operatorname{Attn}\!\left(\operatorname{LN}_{1}^{(\ell)}\!\left(\mathbf{x}^{(\ell)}\right)\right),
\end{equation}
where $\operatorname{Attn}(\cdot)$ is masked self-attention and $\operatorname{LN}_{1}^{(\ell)}$ is the first layer normalization of the block.

The memory branch then normalizes the post-attention states and passes them to the memory cell:
\begin{equation}
\mathbf{d}^{(\ell)} =
\operatorname{MC}^{(\ell)}\!\left(\operatorname{LN}_{2}^{(\ell)}\!\left(\mathbf{h}^{(\ell)}\right);\tau\right),
\end{equation}
where $\operatorname{LN}_{2}^{(\ell)}$ is the second layer normalization, $\operatorname{MC}^{(\ell)}$ is the layer-specific memory cell, and $\tau$ is the routing temperature for soft assignment over centroids. Adding $\mathbf{d}^{(\ell)}$ to the residual stream gives the block output
\begin{equation}
\mathbf{x}^{(\ell+1)} = \mathbf{h}^{(\ell)} + \mathbf{d}^{(\ell)}.
\end{equation}

For each token, the memory cell forms source and target memory states, $\mathbf{c}_{\mathrm{src}}$ and $\mathbf{c}_{\mathrm{tgt}}$, by weighting the layer-normalized centroids with the corresponding soft distributions. The source state represents the token's current location in memory; the target state represents the destination selected through graph propagation and token-conditioned scoring. The returned update is
\begin{equation}
\mathbf{d}^{(\ell)} =
\sigma\!\left(g^{(\ell)}\right)\,
\operatorname{LN}_{\mathrm{disp}}^{(\ell)}
\!\left(\mathbf{c}_{\mathrm{tgt}} - \mathbf{c}_{\mathrm{src}}\right),
\end{equation}
where $g^{(\ell)}$ is a learned scalar gate for layer $\ell$ and $\operatorname{LN}_{\mathrm{disp}}^{(\ell)}$ is a layer normalization applied to the displacement vector.

\paragraph{Memory-branch ablation.}
Here e0 denotes the unmodified trained \ac{GMT} control, not the dense transformer
baseline. Disabling its memory branch and write-back at inference produces pronounced
repetition on the 23-prompt suite
(Section~\ref{sec:generation_ablations}; parentheses give deltas from e0):
\begin{center}
\small
\begin{tabular}{@{}lrr@{}}
\toprule
Condition & Distinct-2 $\uparrow$ & Rep-4 $\downarrow$ \\
\midrule
e0 control & 0.9568 & 0.0213 \\
Memory disabled & 0.0844 ($-0.8724$) & 0.8216 ($+0.8003$) \\
\bottomrule
\end{tabular}
\end{center}
This supports the trained model's dependence on its memory branch; it does not
compare independently trained architectures.

\subsection{Memory Representation}
\label{sec:memory_representation}

For layer $\ell$, the centroid bank is
\begin{equation}
\mathbf{C}^{(\ell)} \in \mathbb{R}^{F \times H}
\end{equation}
where $F$ is the number of memory slots and $H$ is the hidden dimension. Its $i$-th row,
\begin{equation}
\mathbf{c}^{(\ell)}_i \in \mathbb{R}^{H},
\end{equation}
is a learned centroid vector serving as a persistent representational landmark in the layer's memory space. Shared across inputs, the centroids provide stable reference points for positioning token states. Their coordinates are learned jointly with the model and can also change through adaptive write-back.

The centroids are initialized on the unit sphere. During each forward pass, the memory cell first layer-normalizes the raw bank:
\begin{equation}
\widetilde{\mathbf{C}}^{(\ell)} =
\operatorname{LN}_{C}^{(\ell)}\!\left(\mathbf{C}^{(\ell)}\right),
\end{equation}
and then normalizes each row to unit length:
\begin{equation}
\widehat{\mathbf{C}}^{(\ell)}_i =
\frac{\widetilde{\mathbf{c}}^{(\ell)}_i}
{\left\lVert \widetilde{\mathbf{c}}^{(\ell)}_i \right\rVert_2}.
\end{equation}
The direction-normalized centroids $\widehat{\mathbf{C}}^{(\ell)}$ provide a geometrically stable basis for similarity-based source routing. The layer-normalized centroids $\widetilde{\mathbf{C}}^{(\ell)}$ are the vectors used to construct source and target memory states as weighted sums.

Transitions between slots are parameterized by a learned edge-score matrix
\begin{equation}
\mathbf{E}^{(\ell)} \in \mathbb{R}^{F \times F},
\end{equation}
whose entry $E^{(\ell)}_{ij}$ is an unnormalized score expressing the preference for moving from slot $i$ to slot $j$. The graph is directed: in general, $E^{(\ell)}_{ij} \neq E^{(\ell)}_{ji}$. These asymmetric transitions represent directed computational steps between memory regions, not a symmetric similarity relation.

To prevent trivial self-loops from dominating traversal, direct self-edges are excluded by masking the diagonal before converting scores to transition probabilities. With a mask $\mathbf{M} \in \mathbb{R}^{F \times F}$ defined by
\begin{equation}
M_{ij} =
\begin{cases}
0, & i \neq j, \\
-\infty, & i = j,
\end{cases}
\end{equation}
the row-stochastic transition matrix is
\begin{equation}
\mathbf{P}^{(\ell)} =
\operatorname{softmax}\!\left(\mathbf{E}^{(\ell)} + \mathbf{M}\right),
\end{equation}
where the softmax is applied row-wise. Each row of $\mathbf{P}^{(\ell)}$ is therefore a distribution over outgoing transitions to other slots.

\subsection{Source Routing}

Source routing maps the current token state to a distribution over the layer's centroids. Let $\mathbf{z}^{(\ell)}_t \in \mathbb{R}^{H}$ denote the layer-normalized token state entering the memory cell at layer $\ell$ and position $t$:
\begin{equation}
\mathbf{z}^{(\ell)}_t =
\operatorname{LN}_{2}^{(\ell)}\!\left(\mathbf{h}^{(\ell)}_t\right).
\end{equation}
Routing uses direction-normalized versions of both the token state and the centroids:
\begin{equation}
\widehat{\mathbf{z}}^{(\ell)}_t =
\frac{\mathbf{z}^{(\ell)}_t}{\lVert \mathbf{z}^{(\ell)}_t \rVert_2},
\qquad
\widehat{\mathbf{c}}^{(\ell)}_i =
\frac{\widetilde{\mathbf{c}}^{(\ell)}_i}{\lVert \widetilde{\mathbf{c}}^{(\ell)}_i \rVert_2},
\end{equation}
where $\widetilde{\mathbf{c}}^{(\ell)}_i$ is the layer-normalized centroid defined in the previous subsection. Their inner product gives the cosine similarity between token $t$ and centroid $i$:
\begin{equation}
s^{(\ell)}_{t,i} =
\widehat{\mathbf{z}}^{(\ell)}_t \cdot \widehat{\mathbf{c}}^{(\ell)}_i.
\end{equation}
The cosine distance is floored before its reciprocal is used as a routing score:
\begin{equation}
d^{(\ell)}_{t,i} =
\max\!\left(1 - s^{(\ell)}_{t,i}, \varepsilon_{\mathrm{grav}}\right),
\end{equation}
where the small constant $\varepsilon_{\mathrm{grav}} > 0$ avoids numerical instability near unit similarity by preventing division by zero and capping the reciprocal at $1/\varepsilon_{\mathrm{grav}}$.

The source-routing weights use a softmax over temperature-scaled inverse distances rather than the similarities themselves:
\begin{equation}
w^{(\ell)}_{\mathrm{src},\,t,i} =
\frac{
e^{\frac{1}{\tau\, d^{(\ell)}_{t,i}}}
}{
\sum\limits_{j=1}^{F}
e^{\frac{1}{\tau\, d^{(\ell)}_{t,j}}}
},
\end{equation}
where $\tau > 0$ is a temperature parameter. For each token position $t$, this yields a probability distribution over the $F$ centroids:
\begin{equation}
\mathbf{w}^{(\ell)}_{\mathrm{src},\,t} \in \Delta^{F-1}.
\end{equation}
Centroids with smaller distances receive greater weight. Reciprocal distances make the scores rise sharply near alignment, until the distance floor is reached. For fixed distances, higher $\tau$ broadens the source distribution, while lower $\tau$ concentrates it on the closest centroids.

Soft assignment preserves differentiability and represents uncertainty through partial membership across nearby centroids, which is especially important while their geometry is being learned. Write-back instead uses hard source assignments to provide a decisive maintenance signal.

The source weights are propagated through the graph to select a target distribution. They also define the source memory state as a weighted combination of the layer-normalized centroids $\widetilde{\mathbf{c}}^{(\ell)}_i$ (Section~\ref{sec:displacement_readout}).

\subsection{Graph Traversal and Target Selection}

To select a target, \ac{GMT} first propagates the source distribution through the learned transition graph, then combines the result with token-conditioned contextual scores. It does not read directly from the source distribution alone.

Let $\mathbf{w}^{(\ell)}_{\mathrm{src},\,t} \in \Delta^{F-1}$ denote the source distribution for token $t$ at layer $\ell$, and let $\mathbf{P}^{(\ell)} \in \mathbb{R}^{F \times F}$ be the row-stochastic transition matrix defined in Section~\ref{sec:memory_representation}. Graph propagation gives
\begin{equation}
\mathbf{w}^{(\ell)}_{\mathrm{edge},\,t}
=
\mathbf{w}^{(\ell)}_{\mathrm{src},\,t}\,\mathbf{P}^{(\ell)}.
\end{equation}
This operation can be read as a one-step diffusion over the learned centroid graph: it expresses which memory regions become likely after following the transition tendencies associated with the token's source location. By itself, however, graph propagation is unconditional with respect to the specific token content at the current step.

To reintroduce token-specific information, the memory cell uses learned query and key projections at each layer $\ell$:
\begin{equation}
\mathbf{W}^{(\ell)}_{Q} \in \mathbb{R}^{H \times D},
\qquad
\mathbf{W}^{(\ell)}_{K} \in \mathbb{R}^{H \times D},
\end{equation}
where $D$ is the navigation dimension. The query uses the same layer-normalized token state $\mathbf{z}^{(\ell)}_t$ that enters the memory cell:
\begin{equation}
\mathbf{q}^{(\ell)}_t = \mathbf{z}^{(\ell)}_t \mathbf{W}^{(\ell)}_{Q},
\end{equation}
and each centroid key uses its layer-normalized representation:
\begin{equation}
\mathbf{k}^{(\ell)}_i = \widetilde{\mathbf{c}}^{(\ell)}_i \mathbf{W}^{(\ell)}_{K}.
\end{equation}
The contextual compatibility between token $t$ and centroid $i$ is then
\begin{equation}
a^{(\ell)}_{t,i}
=
\frac{\mathbf{q}^{(\ell)}_t \cdot \mathbf{k}^{(\ell)}_i}{\sqrt{D}}.
\end{equation}
These scores are analogous to attention logits \cite{vaswani2017attention}, but evaluate how appropriate each centroid is as a destination for the token's present state. They are not used to aggregate values across sequence positions.

The propagated weights and contextual scores are added before a softmax forms the target distribution:
\begin{equation}
w^{(\ell)}_{\mathrm{tgt},\,t,i}
=
\frac{
e^{w^{(\ell)}_{\mathrm{edge},\,t,i} + a^{(\ell)}_{t,i}}
}{
\sum\limits_{j=1}^{F}
e^{w^{(\ell)}_{\mathrm{edge},\,t,j} + a^{(\ell)}_{t,j}}
}.
\end{equation}
Equivalently, if $\mathbf{a}^{(\ell)}_t \in \mathbb{R}^{F}$ collects the contextual scores over all centroids, then
\begin{equation}
\mathbf{w}^{(\ell)}_{\mathrm{tgt},\,t}
=
\operatorname{softmax}\!\left(
\mathbf{w}^{(\ell)}_{\mathrm{edge},\,t}
+ \mathbf{a}^{(\ell)}_t
\right).
\end{equation}
The target distribution $\mathbf{w}^{(\ell)}_{\mathrm{tgt},\,t}$ lies on the simplex $\Delta^{F-1}$. Whereas $\mathbf{w}^{(\ell)}_{\mathrm{src},\,t}$ estimates where the token currently sits in memory, the target distribution estimates its destination. Graph propagation supplies a structural prior, while token-conditioned scores adapt the destination to the current hidden state and local computational needs, keeping selection from becoming purely graph-driven or purely content-driven. Retaining a distribution over centroids instead of selecting a single node preserves differentiability and allows mixed destinations near boundaries between memory regions.

\subsection{Displacement Readout}
\label{sec:displacement_readout}

The readout converts the source and target distributions into memory states, then returns their displacement in hidden-state space.

Let $\mathbf{w}^{(\ell)}_{\mathrm{src},\,t}, \mathbf{w}^{(\ell)}_{\mathrm{tgt},\,t} \in \Delta^{F-1}$ denote the source and target distributions for token $t$ at layer $\ell$, and let $\widetilde{\mathbf{C}}^{(\ell)} \in \mathbb{R}^{F \times H}$ be the layer-normalized centroid bank. The corresponding source and target memory states are defined by weighted aggregation:
\begin{equation}
\mathbf{c}^{(\ell)}_{\mathrm{src},\,t}
=
\mathbf{w}^{(\ell)}_{\mathrm{src},\,t}\,\widetilde{\mathbf{C}}^{(\ell)},
\end{equation}
\begin{equation}
\mathbf{c}^{(\ell)}_{\mathrm{tgt},\,t}
=
\mathbf{w}^{(\ell)}_{\mathrm{tgt},\,t}\,\widetilde{\mathbf{C}}^{(\ell)}.
\end{equation}
Each state lies in $\mathbb{R}^{H}$ and represents a soft point in the layer's memory space, typically summarizing a local region through a convex combination of multiple centroids.\footnote{Strictly speaking, because the centroids are not constrained to be nonnegative and are layer-normalized before aggregation, the resulting vector is an affine weighted combination rather than a convex combination in the geometric sense. The important point is that the coefficients are nonnegative and sum to one.} It is tied to a single centroid only when the corresponding distribution becomes nearly one-hot.

Subtracting the source memory state from the target memory state gives the memory cell's core readout:
\begin{equation}
\Delta \mathbf{c}^{(\ell)}_t
=
\mathbf{c}^{(\ell)}_{\mathrm{tgt},\,t}
-
\mathbf{c}^{(\ell)}_{\mathrm{src},\,t}.
\end{equation}
The displacement expresses, in hidden-state coordinates, how the token should move according to the layer's memory dynamics. If the source and target states coincide, the displacement is small; if they differ substantially, the memory cell proposes a larger update.

Before adding the displacement to the residual stream, a learned post-processing step applies layer normalization and a scalar gate:
\begin{equation}
\mathbf{d}^{(\ell)}_t
=
\sigma\!\left(g^{(\ell)}\right)\,
\operatorname{LN}^{(\ell)}_{\mathrm{disp}}
\!\left(
\Delta \mathbf{c}^{(\ell)}_t
\right),
\end{equation}
where $g^{(\ell)} \in \mathbb{R}$ is a learned scalar gate and $\operatorname{LN}^{(\ell)}_{\mathrm{disp}}$ is a layer normalization applied to the displacement. The sigmoid bounds the effective scaling while allowing the layer to learn how strongly the memory-cell update influences the residual stream, modulating its contribution instead of forcing it to dominate every token update equally.

In many memory-augmented architectures, readout retrieves relevant memory content and injects it back into the computation \cite{weston2015memory,sukhbaatar2015endtoend,graves2016hybrid}. \ac{GMT} instead returns a directed correction in representation space, not a single stored value. This relational readout is one of its main conceptual departures: the update depends not only on the state reached, but on how it differs from the starting state.

Defining the update as movement between positions aligns with the interpretation of the memory graph as a space of positions and transitions. Returning a vector in the token's hidden dimension also makes the readout compatible with the transformer block's residual structure \cite{vaswani2017attention}, so it can be added directly to the ongoing computation.

\paragraph{Position-specific displacement ablation.}
To remove position specificity, we replace the returned displacement at every position
with its sequence mean, computed after normalization and gating. On the six-prompt
suite (Section~\ref{sec:generation_ablations}), this intervention increases repetition:
\begin{center}
\small
\begin{tabular}{@{}lrr@{}}
\toprule
Condition & Distinct-2 $\uparrow$ & Rep-4 $\downarrow$ \\
\midrule
e0 control & 0.9660 & 0.0065 \\
Mean displacement & 0.5836 ($-0.3824$) & 0.3653 ($+0.3588$) \\
\bottomrule
\end{tabular}
\end{center}
\begin{samepage}
For sample 0 of the speed-of-light prompt, e0 begins with ``the power of the sun and its
ability to be used by the people'', while the intervention produces
``their benefit benefits benefits benefits benefits''. These excerpts illustrate
looping, not factual accuracy. The intervention supports retaining a position-specific
readout in the trained model.

\end{samepage}

\subsection{Memory Write-Back}
\label{sec:writeback}

Write-back complements the displacement readout by updating centroids online during training. Combining explicit reads with explicit memory updates during computation or learning is broadly consistent with the tradition of memory-augmented architectures \cite{graves2014neural,graves2016hybrid}.

Write-back uses the source distribution, not the target distribution: source routing locates the current token in memory and identifies the region to refine. The state written back is the post-memory block output rather than the cell's pre-memory input. Let
\begin{equation}
\mathbf{r}^{(\ell)}_t
=
\mathbf{h}^{(\ell)}_t + \mathbf{d}^{(\ell)}_t
=
\mathbf{x}^{(\ell+1)}_t
\end{equation}
denote that block output at position $t$ in layer $\ell$, and let $\mathbf{w}^{(\ell)}_{\mathrm{src},\,t} \in \Delta^{F-1}$ be the corresponding source-routing distribution. To obtain more decisive centroid updates, write-back converts the forward pass's soft routing weights into hard assignments.

For each token position $t$, the assigned centroid index is
\begin{equation}
i^{(\ell)}_t
=
\arg\max_{1 \leq i \leq F}
w^{(\ell)}_{\mathrm{src},\,t,i}.
\end{equation}
This induces a one-hot assignment vector
\begin{equation}
\mathbf{e}^{(\ell)}_t \in \{0,1\}^{F},
\qquad
e^{(\ell)}_{t,i} =
\mathbb{I}\!\left[i = i^{(\ell)}_t\right].
\end{equation}
These assignments partition the post-block states across all token positions in the current batch and sequence window by their most strongly routed centroid. For the set $\mathcal{T}^{(\ell)}_i$ of positions assigned to centroid $i$, the batch-level update target is their empirical mean:
\begin{equation}
\bar{\mathbf{r}}^{(\ell)}_i
=
\frac{1}{\max\!\left(|\mathcal{T}^{(\ell)}_i|,\varepsilon\right)}
\sum_{t \in \mathcal{T}^{(\ell)}_i}
\mathbf{r}^{(\ell)}_t,
\end{equation}
where $\varepsilon > 0$ is a small constant preventing division by zero in empty-assignment cases.

The centroid update is an exponential moving average with a learned scalar momentum
(Appendix~\ref{app:writeback}). Let
\begin{equation}
m^{(\ell)} = \sigma\!\left(u^{(\ell)}\right),
\end{equation}
where $u^{(\ell)} \in \mathbb{R}$ is a learnable parameter and $\sigma(\cdot)$ is the sigmoid function. The centroid bank is then updated according to
\begin{equation}
\mathbf{c}^{(\ell)}_i
\leftarrow
m^{(\ell)} \mathbf{c}^{(\ell)}_i
+
\left(1 - m^{(\ell)}\right)\bar{\mathbf{r}}^{(\ell)}_i.
\end{equation}
After the update, each centroid is renormalized:
\begin{equation}
\mathbf{c}^{(\ell)}_i
\leftarrow
\frac{\mathbf{c}^{(\ell)}_i}{\lVert \mathbf{c}^{(\ell)}_i \rVert_2}.
\end{equation}
This normalization keeps the centroid magnitudes controlled and makes the geometry of the memory bank more stable over training.

This rule allows the memory representation to evolve on two timescales. Gradients from the main objective and auxiliary losses shape centroids indirectly through end-to-end optimization; write-back directly nudges frequently selected centroids toward their routed states, helping keep these source landmarks aligned with the data regions represented in recent forward passes. The learned momentum controls how conservative or reactive this adaptation is. When $m^{(\ell)}$ is high, centroid updates are slow and memory remains stable; at lower values, the centroids track recent assignments more aggressively.

Write-back is deliberately simple and lightweight: it does not update the transition graph or perform a complex associative memory update at each step. The auxiliary losses and maintenance rules below are used to stabilize this process and prevent memory collapse or underutilization.

\subsection{Auxiliary Objectives}
\label{sec:aux_objectives}

\paragraph{Objective composition.}
Auxiliary losses are integral to training a usable memory structure: they are designed to stabilize centroid geometry, improve memory utilization, and regularize the transition graph. Without them, the model can still optimize the next-token objective, but the learned centroids and edges are more likely to collapse, become redundant, or fail to organize into a meaningful routing structure.

Let $\mathcal{L}_{\mathrm{task}}$ denote the standard next-token cross-entropy loss. Each layer also computes five memory-specific terms: a tracking loss, an orthogonality loss on the centroid bank, a clustering penalty on centroid usage, an edge-entropy regularizer, and an edge-contrastive term that encourages diversity among transition patterns. Their weighted sum is

\begin{equation}
\mathcal{L}_{\mathrm{mem}}
=
\lambda_{\mathrm{track}} \mathcal{L}_{\mathrm{track}}
+
\beta_{\mathrm{ortho}} \mathcal{L}_{\mathrm{ortho}}
+
\lambda_{\mathrm{cluster}} \mathcal{L}_{\mathrm{cluster}}
+
\lambda_{\mathrm{edge}} \mathcal{L}_{\mathrm{edge}}
+
\lambda_{\mathrm{contrast}} \mathcal{L}_{\mathrm{contrast}},
\end{equation}
and the full training objective is
\begin{equation}
\mathcal{L}
=
\mathcal{L}_{\mathrm{task}} + \mathcal{L}_{\mathrm{mem}}.
\end{equation}

\paragraph{Tracking loss.}
\label{sec:tracking_loss}
The tracking loss ensures that the source-routing distribution remains predictive of the hidden state written back into memory. It reconstructs a source-conditioned centroid mixture from the source weights $\mathbf{w}_{\mathrm{src}}$ and the layer-normalized centroid bank $\widetilde{\mathbf{C}}$, then compares it with the post-memory block output $\mathbf{x}^{(\ell+1)}$. With the learned write-back momentum $m = \sigma(u)$, the loss is
\begin{equation}
\mathcal{L}_{\mathrm{track}}
=
\left(1 - m\right)
\operatorname{MSE}
\!\left(
\operatorname{sg}\!\left(\mathbf{x}^{(\ell+1)}\right),
\mathbf{w}_{\mathrm{src}} \widetilde{\mathbf{C}}
\right),
\end{equation}
where $\operatorname{sg}(\cdot)$ denotes stop-gradient and the expression is understood over the batch and sequence dimensions. This replaces a conventional commitment-style penalty \cite{oord2017neural} without forcing a hard match between token states and centroids. The prefactor $(1-m)$ couples the loss to write-back dynamics, reducing tracking pressure when the learned momentum becomes highly conservative.

\paragraph{Centroid orthogonality.}
The orthogonality term regularizes the centroid bank itself. Let $\mathbf{C}_{n}$ denote the row-wise $\ell_{2}$-normalized centroid matrix and let
\begin{equation}
\mathbf{G} = \mathbf{C}_{n}\mathbf{C}_{n}^{\top}
\end{equation}
be its Gram matrix. The orthogonality loss penalizes off-diagonal similarity:
\begin{equation}
\mathcal{L}_{\mathrm{ortho}}
=
\frac{1}{F(F-1)}
\sum_{i \neq j} G_{ij}^{2}.
\end{equation}
The penalty discourages centroid collapse onto nearly identical directions, pushing the bank toward a more distributed set of representational landmarks and helping avoid trivial solutions with many slots concentrated in the same hidden-state region.

\paragraph{Usage clustering.}
The clustering term addresses memory utilization by estimating both the average source usage of centroid $i$ over the current batch of $N$ routed states and the corresponding normalized usage distribution:
\begin{equation}
\bar{u}_i
=
\frac{1}{N}
\sum_{t=1}^{N}
w_{\mathrm{src},t,i},
\qquad
u_i
=
\frac{\bar{u}_i}{\sum_{j=1}^{F}\bar{u}_j + \varepsilon}
\end{equation}
where $u_i$ is the numerically renormalized usage distribution used by the entropy term. The model then forms
\begin{equation}
H(\mathbf{u})
=
-
\sum_{i=1}^{F}
u_i \log(u_i + \varepsilon),
\end{equation}
and the corresponding effective number of active centroids
\begin{equation}
N_{\mathrm{eff}} = e^{H(\mathbf{u})}.
\end{equation}
The clustering penalty is then
\begin{equation}
\mathcal{L}_{\mathrm{cluster}}
=
\max\!\left(
\frac{N_{\mathrm{target}}}{\max(N_{\mathrm{eff}},1)} - 1,\,
0
\right),
\end{equation}
where $N_{\mathrm{target}}$ is a desired lower bound on the effective number of active centroids. This loss becomes zero once centroid usage is sufficiently spread out, but it penalizes situations in which only a small subset of slots carries most of the routing mass.

\paragraph{Edge entropy.}
The edge-entropy term regularizes the row-stochastic transition matrix
\begin{equation}
\mathbf{P} = \operatorname{softmax}(\mathbf{E} + \mathbf{M})
\end{equation}
introduced earlier. For each row $i$, the model computes the outgoing entropy
\begin{equation}
H_i
=
-
\sum_{j=1}^{F}
P_{ij} \log(P_{ij} + \varepsilon).
\end{equation}
The edge-entropy loss then enforces a minimum degree of outgoing uncertainty:
\begin{equation}
\mathcal{L}_{\mathrm{edge}}
=
\frac{1}{F}
\sum_{i=1}^{F}
\max\!\left(H_{\mathrm{target}} - H_i,\,0\right),
\end{equation}
where $H_{\mathrm{target}}$ is a prescribed minimum entropy level. This prevents the transition graph from becoming too sharply peaked too early, which would make traversal brittle and reduce the effective diversity of available moves.

\paragraph{Edge contrast.}
The edge-contrastive term encourages different centroids to induce different outgoing transition patterns. If $\mathbf{P}_{n}$ denotes the row-wise $\ell_{2}$-normalized version of $\mathbf{P}$, then the model forms the row-similarity matrix
\begin{equation}
\mathbf{S} = \mathbf{P}_{n}\mathbf{P}_{n}^{\top}
\end{equation}
and penalizes its off-diagonal mean:
\begin{equation}
\mathcal{L}_{\mathrm{contrast}}
=
\frac{1}{F(F-1)}
\sum_{i \neq j} S_{ij}.
\end{equation}
Minimizing this quantity pushes transition rows away from one another so the graph does not degenerate into centroids with nearly identical outgoing behavior. It complements centroid orthogonality, which regularizes the diversity of memory locations; edge contrast regularizes the diversity of their connections and propagation patterns.

Together, these losses keep routing, centroid geometry, and graph structure aligned with the language-modeling objective.

\subsection{Centroid Maintenance}
\label{sec:centroid_maintenance}

Auxiliary losses shape the memory bank's global geometry but do not directly recycle dead or redundant slots. \ac{GMT} therefore uses lightweight online centroid maintenance during training, separate from per-forward write-back, to address local failures that are difficult to eliminate through gradient-based losses alone.

After each forward pass, the model computes the batch-level source utilization
\begin{equation}
\mathbf{u}^{(\ell)}_{\mathrm{batch}}
=
\frac{1}{N}
\sum_{t=1}^{N}
\mathbf{w}^{(\ell)}_{\mathrm{src},\,t},
\end{equation}
where $N$ is the total number of routed token states in the current batch and $\mathbf{w}^{(\ell)}_{\mathrm{src},\,t} \in \Delta^{F-1}$ is the source-routing distribution. The stored usage statistic is then updated by exponential smoothing:
\begin{equation}
\mathbf{u}^{(\ell)}
\leftarrow
\rho\,\mathbf{u}^{(\ell)}
+
\left(1-\rho\right)\mathbf{u}^{(\ell)}_{\mathrm{batch}},
\end{equation}
where $\rho \in (0,1)$ is a smoothing coefficient. This gives a slowly varying estimate of how often each centroid is being selected over time. In addition to usage, each centroid is assigned an age counter
\begin{equation}
a^{(\ell)}_i \in \mathbb{R}_{\ge 0},
\end{equation}
which is incremented after each write-back step and reset whenever a centroid is reinitialized.

These statistics support dead-centroid reset and similarity-based merging. For reset, centroid $i$ is marked inactive if
\begin{equation}
u^{(\ell)}_i < \delta_{\mathrm{dead}},
\end{equation}
where $\delta_{\mathrm{dead}}$ is a fixed reset threshold. If no centroid satisfies this condition, the reset procedure does nothing. Otherwise, let $\mathcal{D}^{(\ell)}=\{i:u^{(\ell)}_i<\delta_{\mathrm{dead}}\}$ be the set of inactive centroids. Each inactive centroid receives a newly normalized replacement vector
\begin{equation}
\mathbf{r}^{(\ell)}_i
=
\begin{cases}
\operatorname{norm}\!\left(\mathbf{x}^{(\ell)}_{s_i}\right),
& |\mathcal{D}^{(\ell)}| \leq F/2,\\
\operatorname{norm}\!\left(\boldsymbol{\epsilon}_i\right),
\quad
\boldsymbol{\epsilon}_i \sim \mathcal{N}(\mathbf{0},\mathbf{I}),
& |\mathcal{D}^{(\ell)}| > F/2,
\end{cases}
\end{equation}
where $\mathbf{x}^{(\ell)}_{s_i}$ is a sampled current state. Replacements use current-batch geometry when at most half the slots are inactive; otherwise, random unit-sphere directions are used to avoid repopulating many slots from the same narrow batch geometry. The reset rule is
\begin{equation}
\mathbf{c}^{(\ell)}_i \leftarrow \mathbf{r}^{(\ell)}_i,
\qquad
u^{(\ell)}_i \leftarrow \frac{1}{F},
\qquad
a^{(\ell)}_i \leftarrow 0,
\end{equation}
for each centroid $i$ flagged as dead.

The second maintenance rule targets redundancy rather than inactivity. Let
\begin{equation}
\mathbf{C}^{(\ell)}_{n}
\end{equation}
denote the row-wise normalized centroid bank, and let
\begin{equation}
\mathbf{G}^{(\ell)}
=
\mathbf{C}^{(\ell)}_{n}
\bigl(\mathbf{C}^{(\ell)}_{n}\bigr)^{\top}
\end{equation}
be the corresponding similarity matrix. The algorithm searches for centroid pairs $(i,j)$ such that
\begin{equation}
G^{(\ell)}_{ij} > \tau_{\mathrm{merge}},
\end{equation}
where $\tau_{\mathrm{merge}}$ is a similarity threshold. However, this comparison is restricted to mature centroids only. If either centroid is younger than a cooldown threshold,
\begin{equation}
a^{(\ell)}_i < a_{\mathrm{cool}}
\qquad \text{or} \qquad
a^{(\ell)}_j < a_{\mathrm{cool}},
\end{equation}
the pair is ignored. This prevents newly reset centroids from being merged again before they have had time to stabilize, protecting them as they begin to specialize.

When a mature pair exceeds the merge threshold, the model does not average the two centroids. Instead, it keeps the more frequently used one and repurposes the other. If $u_i \ge u_j$, centroid $j$ is selected for replacement; otherwise centroid $i$ is replaced. Denoting the replaced index by $r$, the update takes the form
\begin{equation}
\mathbf{c}^{(\ell)}_r \leftarrow \mathbf{s}^{(\ell)},
\qquad
u^{(\ell)}_r \leftarrow \frac{1}{F},
\qquad
a^{(\ell)}_r \leftarrow 0,
\end{equation}
where $\mathbf{s}^{(\ell)}$ is a normalized replacement vector. Repurposing the redundant slot makes room for a new candidate region of representation space instead of retaining two nearly identical centroids.

Maintenance runs periodically during training, with dead-centroid reset followed by similarity-based merging. These complementary safeguards restore inactive capacity and remove redundancy among active, mature centroids without changing the differentiable source-routing rule used in the forward computation. The local failures they address are plausible, not universal, especially early in training or when the bank has more slots than current routing can use effectively: insufficient probability mass can prevent some centroids from specializing, while others remain active but drift toward nearly redundant directions that waste capacity.

\section{Experimental Setup}
\label{sec:experiments}

\subsection{Base Model Instantiation}

The base v7 study uses a decoder-only language model \cite{brown2020language} with
\begin{equation}
L = 16
\end{equation}
transformer blocks, hidden dimension
\begin{equation}
H = 768,
\end{equation}
and
\begin{equation}
N_{h} = 12
\end{equation}
attention heads per layer. The corresponding per-head dimension is therefore
\begin{equation}
d_{h} = \frac{H}{N_{h}} = 64.
\end{equation}
The maximum sequence length is
\begin{equation}
T_{\max} = 1024,
\end{equation}
and the vocabulary size used by the model head is
\begin{equation}
V = 50{,}257.
\end{equation}
With these settings, the instantiated base model contains
\begin{equation}
82{,}213{,}152
\end{equation}
trainable parameters.

Each transformer block contains one memory cell in place of the usual \ac{FFN}. The memory bank of each cell has
\begin{equation}
F = 128
\end{equation}
centroids, and the query-key navigation subspace has dimension
\begin{equation}
D = 128.
\end{equation}
Because the centroid bank is block-specific rather than shared across layers, the model contains a total of
\begin{equation}
L \cdot F = 16 \times 128 = 2048
\end{equation}
learned centroid slots across the full network. Each layer can organize its memory differently; no global centroid vocabulary is imposed across depth.

Centroid banks are initialized by projecting Gaussian samples onto the unit sphere. Each memory cell's learned scalar output-gate and write-back-momentum parameters are initialized at $1.0$ and $4.6$, giving initial sigmoid values of approximately $0.731$ and $0.99$, respectively. Token and positional embeddings are randomly initialized and remain trainable throughout, with no pretrained initialization or freeze stage. After the final layer normalization, the output projection is tied to the token embedding matrix \cite{press2017using}.

The model applies dropout with rate
\begin{equation}
p_{\mathrm{embed}} = 0.1
\end{equation}
to the summed input embeddings and
\begin{equation}
p_{\mathrm{attn}} = 0.1
\end{equation}
within causal self-attention. No separate dropout is introduced inside the memory cell itself beyond the normalization and gating operations described earlier.

The routing mechanism is instantiated with minimum routing distance
\begin{equation}
\varepsilon_{\mathrm{grav}} = 0.01
\end{equation}
and with a monotone logarithmic temperature schedule from
\begin{equation}
\tau_{\max} = 1.0
\qquad \text{to} \qquad
\tau_{\min} = 0.1.
\end{equation}
The memory objective combines the five auxiliary terms with
\begin{equation}
\lambda_{\mathrm{track}} = 1.0,\quad
\beta_{\mathrm{ortho}} = 0.05,\quad
\lambda_{\mathrm{cluster}} = 0.3,\quad
\lambda_{\mathrm{edge}} = 0.1,\quad
\lambda_{\mathrm{contrast}} = 0.5,
\end{equation}
and uses
\begin{equation}
N_{\mathrm{target}} = \frac{F}{4},
H_{\mathrm{target}} = 4.0,
\qquad
\delta_{\mathrm{dead}} = 10^{-3},
\end{equation}
\begin{equation}
\tau_{\mathrm{merge}} = 0.95,
\qquad
a_{\mathrm{cool}} = 100.
\end{equation}
These values define the base \ac{GMT} configuration and are not claimed to be uniquely optimal. The
moderate scale supports nontrivial attention--memory interactions while allowing the
memory mechanism to be examined without conflating it with very-large-scale model
effects. It is intended as a faithful base for studying \ac{GMT}'s architectural behavior,
not a maximally scaled instance.

\subsection{Datasets, Tokenization, and Splits}

\subsubsection{OpenWebText Preparation}

OpenWebText \cite{gokaslan2019openwebtext} is prepared as a flat autoregressive token stream.
Documents are tokenized with the GPT-2 tokenizer \cite{radford2019language}; an end-of-text token is
appended after each document, and the resulting token ids are stored as contiguous unsigned
16-bit integer arrays. The training and validation streams are then memory-mapped during
training rather than loaded fully into memory.

The default preparation cap is
\begin{equation}
N_{\mathrm{train,max}} = 3 \times 10^{9}
\end{equation}
training tokens. If a positive cap is used, the validation stream is capped at one tenth of
the training cap; if the cap is disabled, all available tokens from the corresponding
document ranges are retained. No document-level filtering is applied beyond the upstream
dataset, and no separators are added other than the tokenizer's end-of-text marker,
keeping the stream close to the standard GPT-style autoregressive setup.

\subsubsection{Tokenization and Sequence Packing}

Let
\begin{equation}
\mathbf{s} = (s_0, s_1, \ldots, s_{N-1})
\end{equation}
denote the concatenated token stream after tokenization and end-of-text insertion. The
training dataset forms fixed-length, non-overlapping windows of length
$T_{\max}=1024$. For sequence index $i$, the model input and target are
\begin{equation}
\mathbf{x}^{(i)} = (s_{iT_{\max}}, \ldots, s_{iT_{\max}+T_{\max}-1}),
\end{equation}
\begin{equation}
\mathbf{y}^{(i)} = (s_{iT_{\max}+1}, \ldots, s_{iT_{\max}+T_{\max}}).
\end{equation}
Each example reads $T_{\max}+1$ consecutive tokens from the memory-mapped stream:
the first $T_{\max}$ tokens form the input, and the last $T_{\max}$ tokens form the
one-token-shifted target.

The number of usable examples in a shard of length $N$ is
\begin{equation}
N_{\mathrm{seq}} = \left\lfloor \frac{N-1}{T_{\max}} \right\rfloor .
\end{equation}
Sequence boundaries follow fixed stream offsets, not document boundaries,
so the decoder-only language model trains on a continuous stream with documents separated
only by end-of-text tokens.

\subsubsection{Training and Validation Split}

The train-validation split is performed before tokenization at the document-index level.
Let $D$ be the number of OpenWebText training documents. The first
\begin{equation}
\left\lfloor 0.95 D \right\rfloor
\end{equation}
documents are assigned to the training stream, and the remaining documents are assigned to
the validation stream. The split is deterministic with respect to the dataset order. During
training, packed training windows are shuffled, while validation windows are evaluated in a
fixed order.

The validation set is used for two distinct purposes: estimating next-token predictive loss
and selecting the lowest-validation-loss checkpoint. It is not used to tune the architecture
during a run, and no validation examples are mixed into the training stream.

\subsection{Training Protocol}

\subsubsection{Optimization and Learning-Rate Schedule}

The base \ac{GMT} v7 model is trained as a standard autoregressive language model with
teacher forcing. For a minibatch of size $B$ and sequence length $T$, the next-token loss is
\begin{equation}
\mathcal{L}_{\mathrm{LM}}
= -\frac{1}{BT}
\sum_{b=1}^{B}
\sum_{t=1}^{T}
\log p_{\theta}\!\left(y_{b,t}\mid x_{b,\leq t}\right).
\end{equation}
The objective optimized during training adds the memory-specific auxiliary terms defined in
Section~\ref{sec:aux_objectives}:
\begin{equation}
\begin{aligned}
\mathcal{L}_{\mathrm{train}}
&= \mathcal{L}_{\mathrm{LM}}
+ \lambda_{\mathrm{track}}\mathcal{L}_{\mathrm{track}}
+ \beta_{\mathrm{ortho}}\mathcal{L}_{\mathrm{ortho}}
\\
&\quad
+ \lambda_{\mathrm{cluster}}\mathcal{L}_{\mathrm{cluster}}
+ \lambda_{\mathrm{edge}}\mathcal{L}_{\mathrm{edge}}
+ \lambda_{\mathrm{contrast}}\mathcal{L}_{\mathrm{contrast}}.
\end{aligned}
\end{equation}

Optimization uses AdamW with learning rate $3\times10^{-4}$, weight decay $0.1$, and
coefficients $(\beta_1,\beta_2)=(0.9,0.95)$. The scheduler is cosine decay with
2,000 warmup steps. The physical minibatch size is
\begin{equation}
B=8,
\end{equation}
with gradient accumulation over
\begin{equation}
A=33
\end{equation}
minibatches. Thus each optimizer update sees
\begin{equation}
B \cdot A \cdot T_{\max}
= 8 \times 33 \times 1024
= 270{,}336
\end{equation}
tokens, ignoring any final incomplete accumulation window at the end of an epoch. Gradients
are clipped to norm $1.0$. Training uses bfloat16 mixed precision. The base run is
configured for two epochs over the prepared stream.

\subsubsection{Routing-Temperature Schedule}

Routing temperature follows a deterministic schedule indexed by the global optimizer step.
If $s$ is the current optimizer step and $S$ is the
planned number of optimizer steps, define
\begin{equation}
p(s) = \min\left(\frac{s}{S}, 1\right)
\end{equation}
and
\begin{equation}
\rho(s) = \log\left(1 + (\mathrm{e}-1)p(s)\right).
\end{equation}
The temperature used by all memory cells at step $s$ is then
\begin{equation}
\tau(s)
= \tau_{\max}
\cdot
\left(\frac{\tau_{\min}}{\tau_{\max}}\right)^{\rho(s)},
\end{equation}
with $\tau_{\max}=1.0$ and $\tau_{\min}=0.1$. In the base configuration, this simplifies to
\begin{equation}
\tau(s)=0.1^{\rho(s)}.
\end{equation}
The schedule starts with softer gravitational source routing because overly sharp
assignments would make the bank brittle while centroids and edges are still unorganized.
As training proceeds, sharper routing encourages more decisive centroid usage.

\subsubsection{Memory Maintenance During Training}
\label{sec:training_maintenance}

Online write-back uses source routing $w_{\mathrm{src}}$ to update centroids on every
forward pass (Section~\ref{sec:writeback}), separately from optimizer-step maintenance.

Periodic maintenance runs after optimizer steps. Every
\begin{equation}
K_{\mathrm{maint}}=110
\end{equation}
global optimizer steps, each block first resets dead centroids and then repurposes one
centroid from each overly similar mature pair. Dead centroids are detected by usage below
$10^{-3}$. Similarity-based merging uses cosine similarity threshold $0.95$ and excludes
centroids younger than 100 maintenance-age units. These safeguards are used to prevent
unused or duplicated slots from consuming the base v7 memory budget; they are not intended
as a general memory-allocation algorithm.

\subsubsection{Validation and Checkpointing}

Validation computes the same next-token cross-entropy as training, but without auxiliary
memory losses in the reported scalar validation value. Let $\mathcal{B}_{\mathrm{val}}$
denote the ordered validation minibatches and let
\[
K_{\mathrm{val}}=\min(512,|\mathcal{B}_{\mathrm{val}}|).
\]
The reported validation loss is
\begin{equation}
\mathcal{L}_{\mathrm{val}}
=
\frac{1}{K_{\mathrm{val}}}
\sum_{k=1}^{K_{\mathrm{val}}}
\left[
-\frac{1}{B_kT}
\sum_{b=1}^{B_k}
\sum_{t=1}^{T}
\log p_{\theta}\!\left(y^{(k)}_{b,t}\mid x^{(k)}_{b,\leq t}\right)
\right],
\end{equation}
and the reported perplexity is
\begin{equation}
\mathrm{PPL} = e^{\mathcal{L}_{\mathrm{val}}},
\end{equation}
with the exponent clipped in the numerical evaluation only to avoid overflow. Checkpoint
selection is based on the lowest validation loss, with periodic recovery checkpoints retained
during training.

Source-usage statistics and centroid write-back remain active during validation, following
the same adaptive forward dynamics as training. Validation therefore measures the base v7
model with adaptive memory, not frozen memory. Comparisons with models whose parameters
remain unchanged during evaluation must account for this protocol distinction.

Alongside validation loss and perplexity, training records memory diagnostics used for
analysis in Section~\ref{sec:results}. These include mean and minimum effective centroid
count, total dead centroids, centroid cosine similarity, coverage, gate statistics,
write-back momentum statistics, edge entropy, maximum edge mass, and edge-row similarity.

\subsection{Baselines and Evaluation Metrics}
\label{sec:evaluation}

\subsubsection{Dense GPT Baseline}

The comparison baseline is a decoder-only transformer with the same outer language-modeling
setup: vocabulary size $50{,}257$, context length 1024, 16 blocks, hidden dimension 768, 12
attention heads, tied input-output embeddings, learned positional embeddings, and the same
dropout rates. The architectural difference is confined to the second sublayer of each
block. Instead of a memory cell, the baseline uses a conventional two-layer \ac{MLP}
\begin{equation}
\operatorname{MLP}(h)
= W_2\,\operatorname{GELU}(W_1 h),
\end{equation}
with intermediate width
\begin{equation}
H_{\mathrm{ff}}=1050.
\end{equation}
This width gives approximately 103M trainable parameters, more than the 82.2M in base
\ac{GMT} v7. The dense model also shares the tokenizer and training pipeline. It is a
strong architectural baseline, not a parameter-matched control.

\subsubsection{Validation-Loss Evaluation}

Validation loss measures next-token predictive fit on held-out OpenWebText. Comparisons
with the dense baseline require the cross-entropy, derived perplexity, optimizer step,
and amount of training data consumed, together with the checkpoint and adaptive evaluation
protocol described above. The scalar loss alone does not specify the comparison.

\subsubsection{ARC-Easy Evaluation}
\label{sec:arc_easy_protocol}

The standalone ARC-Easy evaluation follows a multiple-choice log-likelihood protocol for the
AI2 Reasoning Challenge \cite{clark2018think}. For a question $q$ and answer choices
$a_1,\ldots,a_K$, the evaluator forms the textual context
\begin{equation}
c(q)=\operatorname{concat}\!\left(
\texttt{Question:\ }, q, \langle\mathrm{newline}\rangle, \texttt{Answer:}
\right)
\end{equation}
and scores each answer as a continuation under the language model. If
$a_k=(a_{k,1},\ldots,a_{k,m_k})$ is the tokenized continuation, its raw score is
\begin{equation}
S_{\mathrm{raw}}(a_k)
= \sum_{j=1}^{m_k}
\log p_{\theta}\!\left(a_{k,j}\mid c(q), a_{k,<j}\right).
\end{equation}
The length-normalized score is
\begin{equation}
S_{\mathrm{norm}}(a_k)
= \frac{1}{m_k} S_{\mathrm{raw}}(a_k).
\end{equation}
The predicted raw and normalized answers are
\begin{equation}
\hat{k}_{\mathrm{raw}} = \arg\max_k S_{\mathrm{raw}}(a_k),
\qquad
\hat{k}_{\mathrm{norm}} = \arg\max_k S_{\mathrm{norm}}(a_k).
\end{equation}
The evaluator reports both raw accuracy and length-normalized accuracy. This avoids
treating a single scalar accuracy as definitive when answer-choice lengths differ.

\subsubsection{Multi-Benchmark Zero-Shot Evaluation}

For broader evaluation, we adopt the standardized language-model evaluation protocol of
Gao et al. \cite{gao2024framework}. Scores for the base v7 model are computed through the
standard log-likelihood and rolling-log-likelihood interfaces, using the same GPT-2
tokenization and the model's native context window. In the zero-shot setting,
\begin{equation}
n_{\mathrm{fewshot}}=0,
\end{equation}
so each benchmark example is evaluated without task-specific demonstrations. Result tables
therefore identify the task set, shot count, checkpoint, tokenizer, context length, and
batch size used for evaluation.

This standardized evaluation tests whether the graph-memory replacement preserves enough
language-modeling and commonsense behavior to remain competitive with the dense transformer
baseline under the same checkpointing and tokenization conventions. It is not intended to
establish broad state-of-the-art performance.

\subsubsection{Selected Generation Ablations}
\label{sec:generation_ablations}
The selected diagnostic ablations use saved completions. Memory disabling and
orthogonality removal use 23 prompts with ten samples each; mean displacement uses six
prompts with three samples each. Decoding uses temperature 0.1, top-$k$ 50, top-$p$ 0.92,
repetition penalty 1.3, and at most 120 new tokens, with seeds $1337+s$ for sample $s$.
Each completion starts from the checkpoint state; enabled memory uses adaptive write-back
and the checkpoint routing temperature.

Distinct-$n$ is the fraction of unique whitespace-token $n$-grams; Rep-4 is the fraction
of four-gram occurrences repeating an earlier occurrence. Scores are averaged over
completions, with $\Delta=\text{ablation}-\text{e0}$ relative to the full-GMT control
for the respective suite. These are lexical diagnostics, not semantic-quality scores or
significance tests. The independently trained e02 control contextualizes checkpoint
variation.

\section{Results}
\label{sec:results}

\subsection{Predictive Performance}

We compare the base v7 \ac{GMT} model with the dense GPT-style baseline on held-out
OpenWebText loss, standalone ARC-Easy accuracy, and zero-shot benchmarks. The results are
interpreted within the scope of this configuration, accounting for the parameter-count
difference.

\subsubsection{Validation-Loss Comparison Against the Baseline}

Table~\ref{tab:val_loss} reports the validation cross-entropy and derived perplexity for
both models at their respective best checkpoints, selected by the lowest validation loss
observed during training on the OpenWebText stream.

\begin{table}[htbp]
	\centering
	\caption{Validation-loss comparison at the best checkpoint.}
	\label{tab:val_loss}
	\begin{tabular}{lrrrrr}
		\toprule
		Model & Params & Step & Val.\ Loss & PPL \\
		\midrule
		Baseline GPT-2 & 103.0M & 22{,}096 & 3.2903 & 26.85 \\
		\ac{GMT} v7     &  82.2M & 18{,}310 & 3.5995 & 36.58 \\
		\bottomrule
	\end{tabular}
\end{table}

The baseline reaches a lower validation loss (3.2903 vs.\ 3.5995, a gap of $\Delta = 0.309$
nats) and perplexity (26.85 vs.\ 36.58). This is expected: its 20.8M additional parameters
are concentrated in the 16$\times$ MLP FFN blocks that \ac{GMT} replaces with graph-memory
cells. Because the models are matched in block count rather than parameter count, the
baseline provides an upper bound on what a standard dense transformer achieves at this
depth and hidden size, not a parameter-matched control. Validation loss and perplexity
therefore primarily indicate training quality and relative predictive fit, not same-capacity
efficiency. Despite its 20\% parameter disadvantage, \ac{GMT} reaches a stable training
trajectory and a well-formed validation-loss curve, confirming that the graph-memory
mechanism does not impede convergence.

\subsubsection{Standalone ARC-Easy Evaluation}

Table~\ref{tab:arc_easy_standalone} reports the standalone ARC-Easy test-split
results (2{,}376 examples, 0-shot), scored independently of the
lm-evaluation-harness framework using the raw and length-normalized
log-likelihood protocol described in Section~\ref{sec:arc_easy_protocol}.

\begin{table}[htbp]
	\centering
	\caption{ARC-Easy test-split accuracy (0-shot, 2{,}376 examples).}
	\label{tab:arc_easy_standalone}
	\begin{tabular}{lcc}
		\toprule
		Model & Acc (raw) & Acc (norm) \\
		\midrule
		Baseline GPT-2 & \textbf{0.3893} & \textbf{0.3561} \\
		\ac{GMT} v7     & 0.3704 & 0.3384 \\
		\midrule
		$\Delta$ & $-$0.019 & $-$0.018 \\
		\bottomrule
	\end{tabular}
\end{table}

Despite the 20.8M-parameter difference, \ac{GMT} trails the larger baseline by 1.9 percentage
points on raw accuracy and 1.8 points on length-normalized accuracy. Both gaps are consistent
in sign with the validation-loss advantage: the model with greater FFN capacity assigns
higher likelihood to the correct answer in the majority of cases where the models disagree.

The absolute accuracy of both models ($\sim$37--39\%) reflects the difficulty of zero-shot
ARC-Easy at this scale under standard language-modeling pretraining and is not intended
to match fine-tuned or instruction-tuned results. Parameter-matched comparisons are left
for future work.

\subsubsection{Multi-Benchmark Zero-Shot Evaluation}

Table~\ref{tab:lmeval} reports zero-shot results across four benchmarks evaluated with
the \texttt{lm-evaluation-harness} framework \cite{gao2024framework}: ARC-Easy
\cite{clark2018think}, HellaSwag \cite{zellers2019hellaswag}, PIQA
\cite{bisk2020piqa}, and WinoGrande \cite{sakaguchi2021winogrande}.

\begin{table}[htbp]
	\centering
	\caption{Zero-shot multi-benchmark evaluation (lm-evaluation-harness, 0-shot).}
	\label{tab:lmeval}
	\begin{tabular}{llccc}
		\toprule
		Task & Metric & Baseline & \ac{GMT} v7 & $\Delta$ \\
		\midrule
		\multirow{2}{*}{ARC-Easy}  & acc      & \textbf{0.3893} & 0.3746 & $-$0.015 \\
		& acc\_norm & \textbf{0.3569} & 0.3476 & $-$0.009 \\
		\multirow{2}{*}{HellaSwag} & acc      & \textbf{0.2693} & 0.2665 & $-$0.003 \\
		& acc\_norm & \textbf{0.2750} & 0.2734 & $-$0.002 \\
		\multirow{2}{*}{PIQA}      & acc      & \textbf{0.5947} & 0.5778 & $-$0.017 \\
		& acc\_norm & \textbf{0.5958} & 0.5789 & $-$0.017 \\
		WinoGrande                 & acc      & 0.5051 & \textbf{0.5146} & $+$0.010 \\
		\bottomrule
	\end{tabular}
\end{table}

On the six accuracy-based metrics where the baseline leads, gaps range from 0.2 to
1.7 percentage points, consistent with \ac{GMT} v7's parameter and validation-loss
disadvantage. HellaSwag has the smallest deficit ($<$0.3 pp on both metrics), suggesting
that sentence-completion tasks requiring multi-step contextual coherence are less sensitive
to the FFN-to-memory substitution than answer-selection tasks such as PIQA.

WinoGrande is the exception: \ac{GMT} v7 \emph{outperforms} the baseline by
1.0 percentage point (0.5146 vs.\ 0.5051). The task tests pronoun resolution requiring
implicit commonsense reasoning \cite{sakaguchi2021winogrande}. We hypothesize that
graph-memory routing, which dynamically indexes and retrieves structured associations,
is particularly suited to this referent-disambiguation pattern. The result comes from a
single evaluation run on a single checkpoint and is preliminary, but motivates targeted
analysis of which reasoning sub-types benefit from persistent relational memory.

\subsection{Memory Dynamics and Utilization}

In the base v7 run, we track centroid stability, slot utilization, routing structure,
and edge diversity alongside predictive loss to assess memory organization during
training and evaluation. A graph-memory block can match language-modeling behavior only if its
memory bank avoids collapse, remains sufficiently active, and develops interpretable
transition patterns.

\subsubsection{Centroid Stability}

Three failure modes motivated the base v7 maintenance design and guide these
diagnostics:

\begin{itemize}
	\item \textbf{Collapse to a single attractor}: routing mass concentrates on one
	or a few centroids, giving the model a trivial memory solution rather than a distributed graph state.
	Persistent collapse during early training was treated as a failed development configuration, so the final diagnostics track whether slot usage remains distributed.
	\item \textbf{Centroid under-utilization}: some slots can receive negligible assignment mass, either as a short early-training transient or as a persistent dead-centroid state.
	The maintenance procedure resets long-inactive slots and merges near-duplicate centroids; the relevant question is whether such corrections fade as the memory geometry stabilizes.
	\item \textbf{Momentum-related instability}: write-back momentum controls how quickly centroids follow assigned post-memory states.
	Lower momentum increases adaptability but can destabilize early centroid geometry, whereas high momentum improves stability at the cost of slower adaptation.
	The base v7 run starts in the high-momentum regime and learns a slightly larger value over training; this is descriptive evidence, not a controlled momentum ablation.
\end{itemize}

\subsubsection{Routing Utilization}

We visualize source-to-target centroid transitions for three short text probes---narrative,
political, and scientific---to inspect whether the learned memory graph exposes coherent
routing across linguistic contexts. These are qualitative diagnostics, not separate
benchmarks.
Representative excerpts are:

\begin{itemize}
	\item \textbf{Narrative}: "Maria had lived in the crumbling house at the edge of the forest for twenty years before the letter arrived. The envelope was yellowed, its wax [...]"
	\item \textbf{Politics}: "The central bank raised interest rates by 25 basis points on Thursday, its seventh consecutive increase, citing persistent inflationary pressure across [...]"
	\item \textbf{Science}: "The electron was first identified by J. J. Thomson in 1897 during experiments with cathode rays. His measurements of the charge-to-mass ratio [...]"
\end{itemize}

\begin{figure}[htbp]
	\centering
	\includegraphics[width=\textwidth]{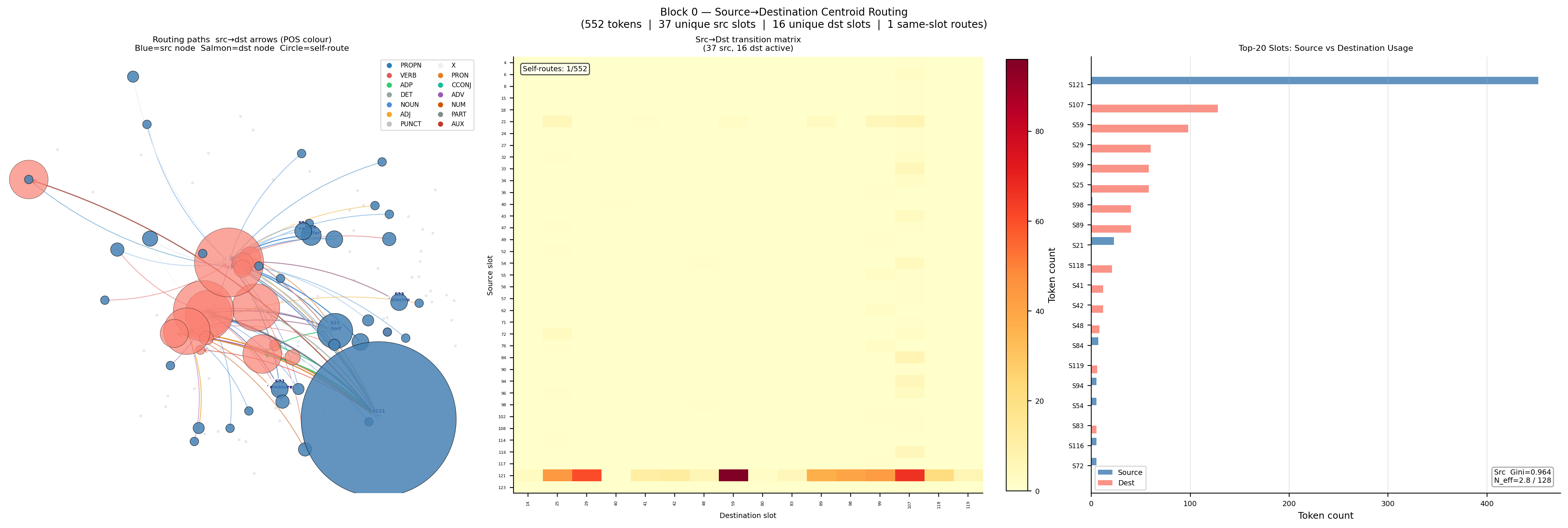}
	\caption{Slot-routing flow at Block~00.}
	\label{fig:slot_routing_flow_block00}
\end{figure}

\begin{figure}[htbp]
	\centering
	\includegraphics[width=\textwidth]{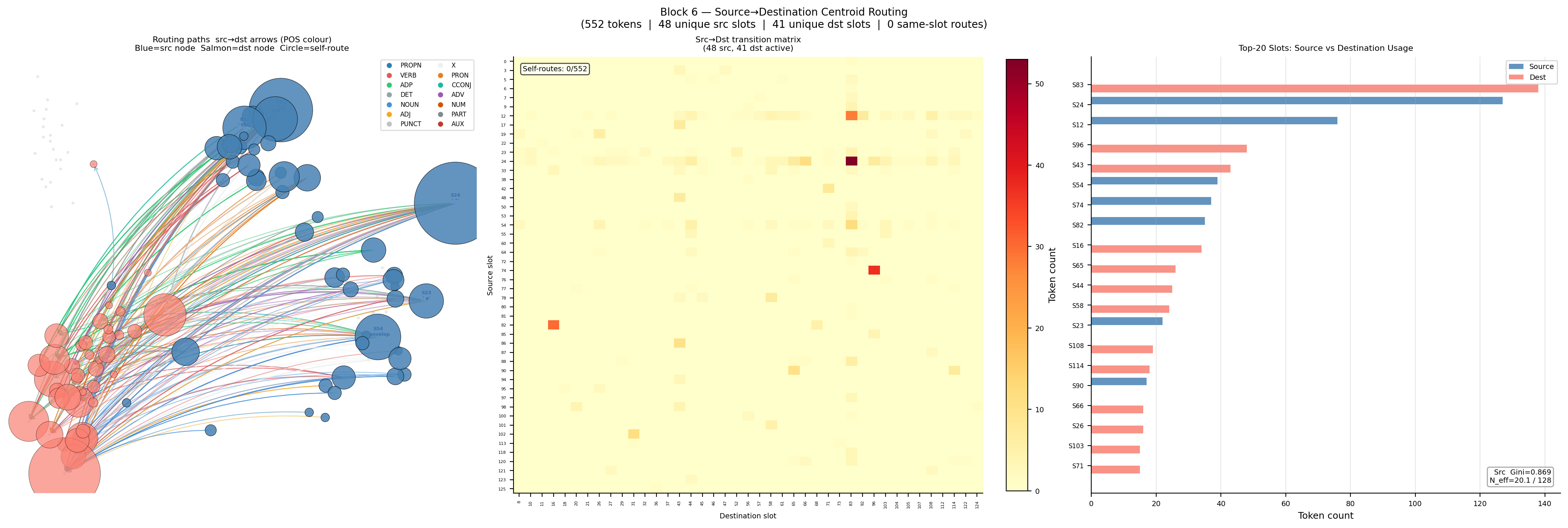}
	\caption{Slot-routing flow at Block~06.}
	\label{fig:slot_routing_flow_block06}
\end{figure}

\begin{figure}[htbp]
	\centering
	\includegraphics[width=\textwidth]{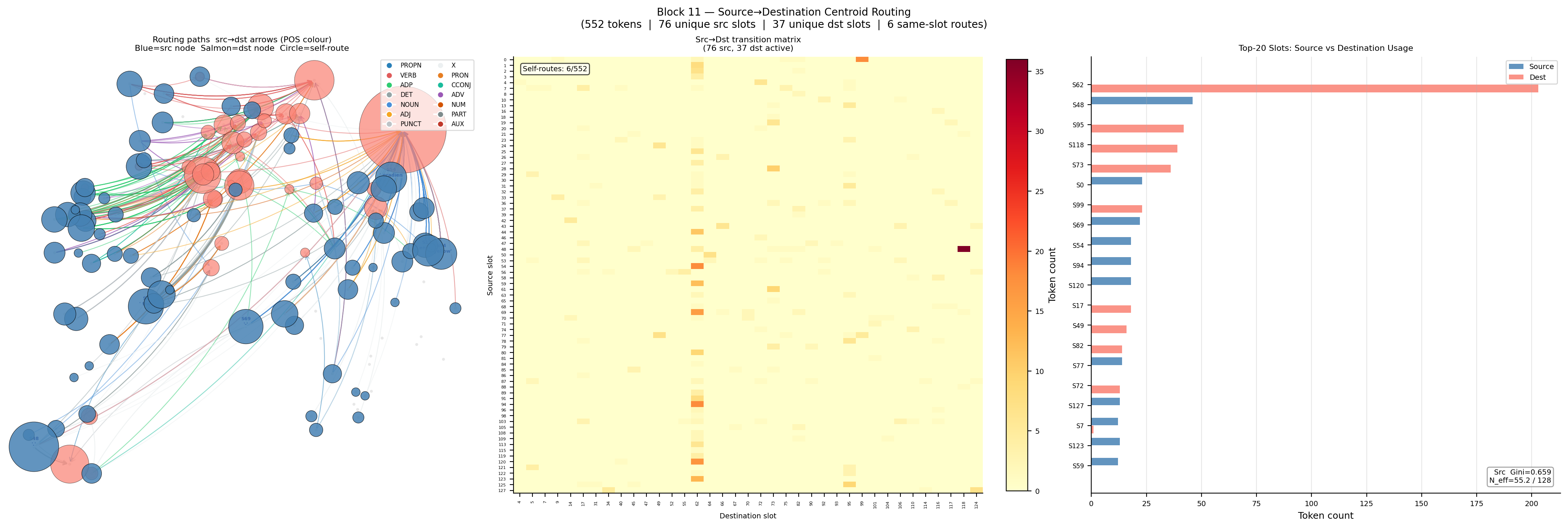}
	\caption{Slot-routing flow at Block~11.}
	\label{fig:slot_routing_flow_block11}
\end{figure}

\begin{figure}[htbp]
	\centering
	\includegraphics[width=\textwidth]{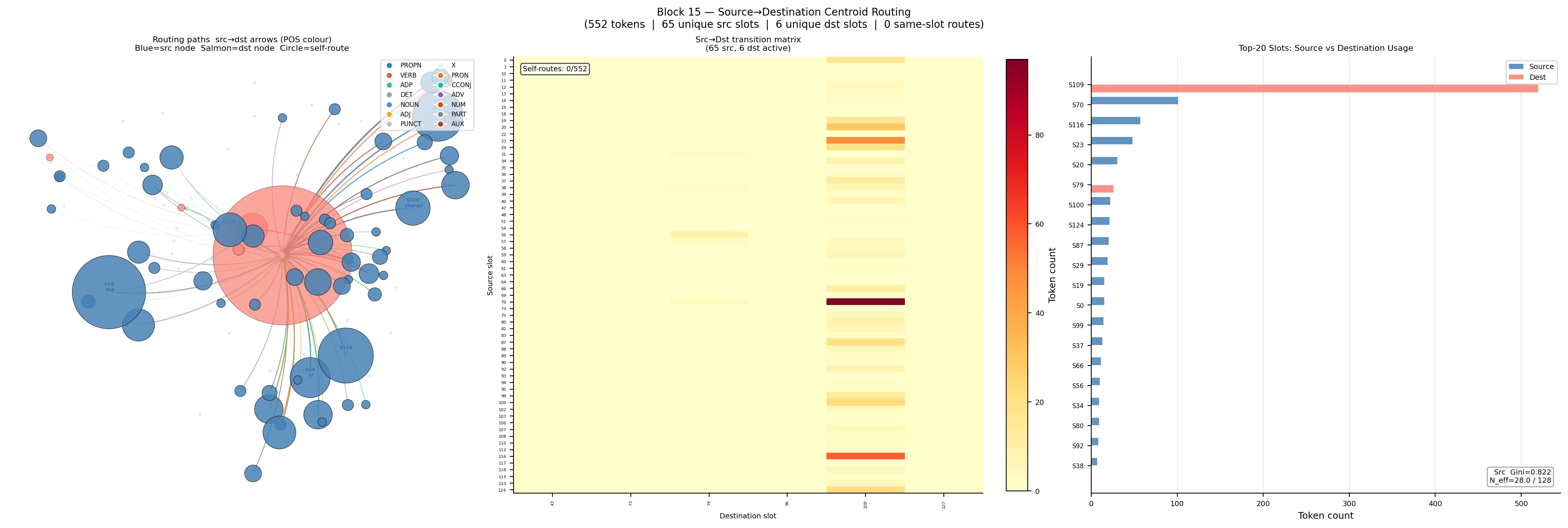}
	\caption{Slot-routing flow at Block~15. The sequence from Blocks~00, 06, 11, and 15 shows the transition from early intake, through mid-depth routing diversification, to late-stage output compression.}
	\label{fig:slot_routing_flow_block15}
\end{figure}

For each probe, routing flows at four memory blocks are shown in
Figures~\ref{fig:slot_routing_flow_block00}, \ref{fig:slot_routing_flow_block06},
\ref{fig:slot_routing_flow_block11}, and~\ref{fig:slot_routing_flow_block15}.
These views are structural diagnostics, not proof of semantic disentanglement.

\begin{figure}[htbp]
	\centering
\includegraphics[width=\textwidth]{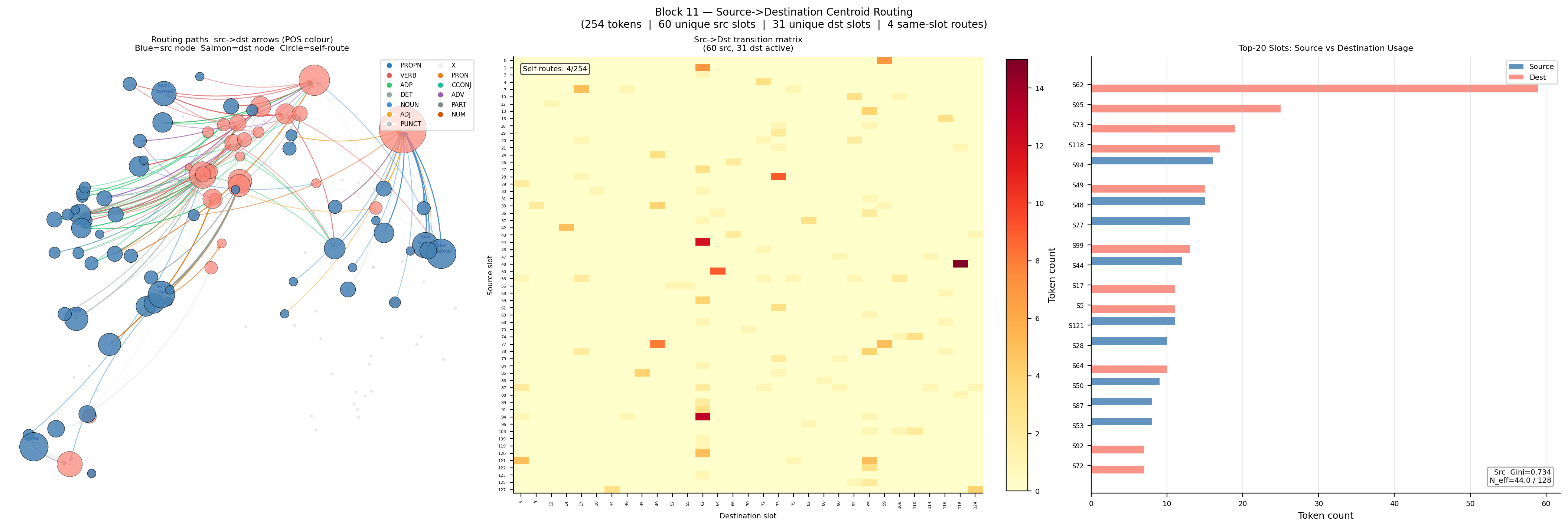}\\[0.5em]
\includegraphics[width=\textwidth]{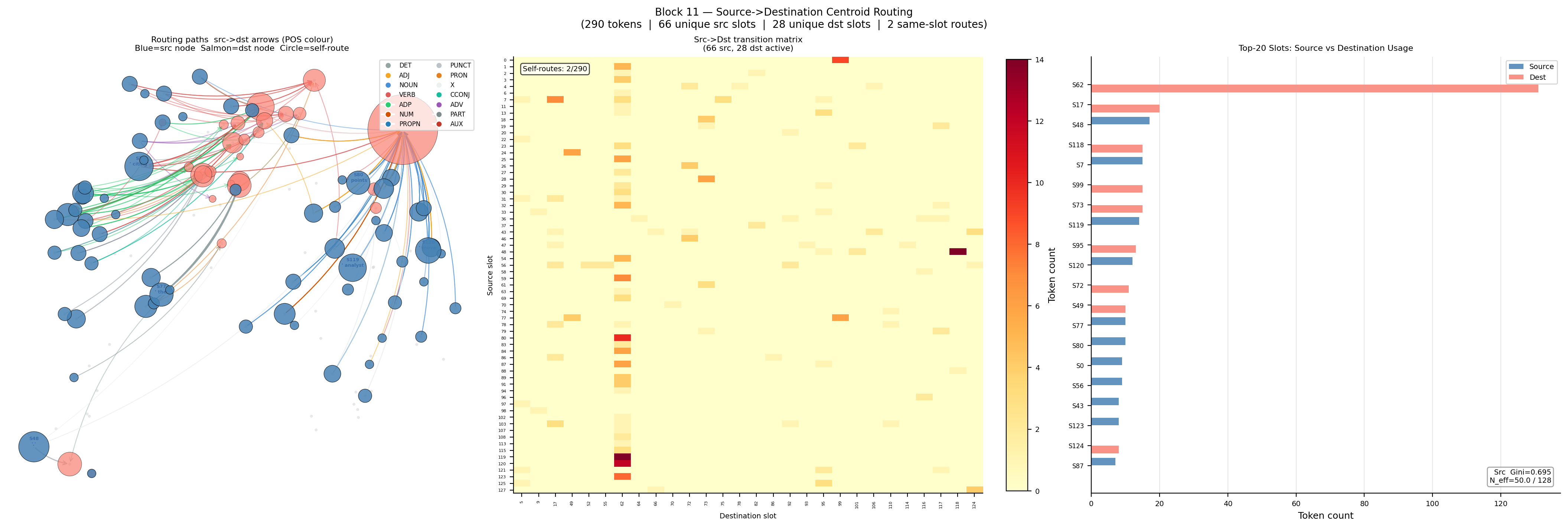}\\[0.5em]
\includegraphics[width=\textwidth]{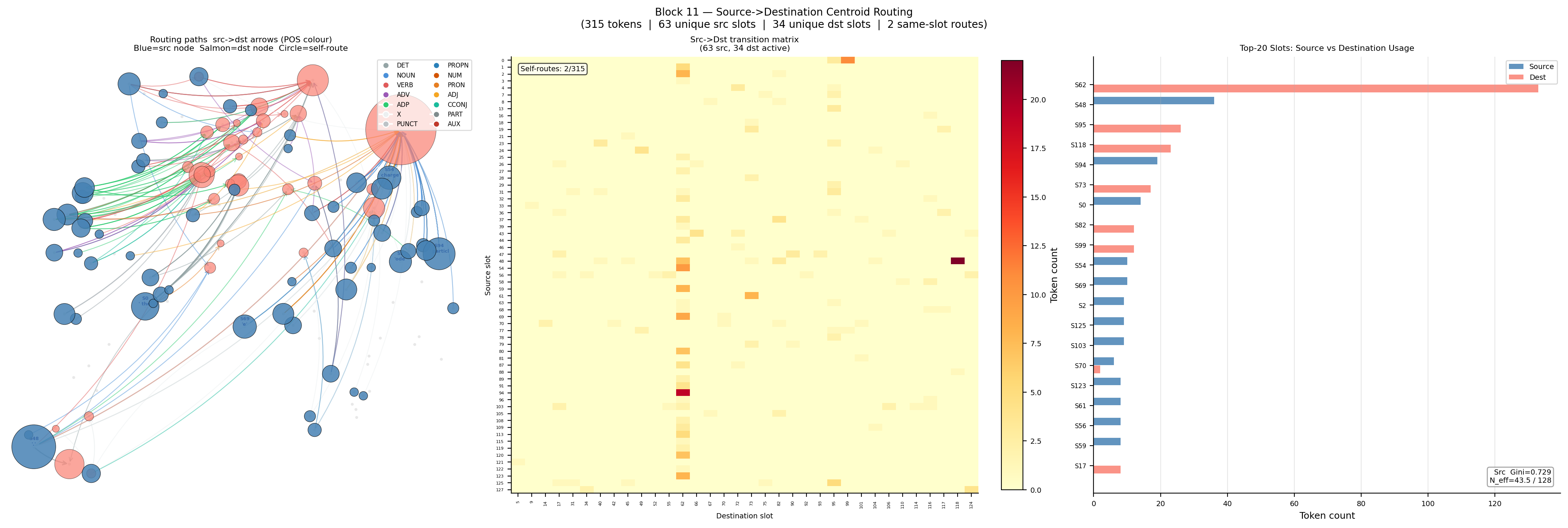}
	\caption{Topic-separated Block~11 routing flows for the narrative, political, and scientific probes. The comparison preserves the same visual encoding across topics and highlights how a shared block-level routing regime can coexist with topic-dependent centroid activation.}
	\label{fig:block11_topic_routing_comparison}
\end{figure}

The part-of-speech (POS) paths make routing structurally inspectable: they show where
grammatical token groups move through the memory graph and how those movements vary
with discourse context, not only local grammatical role. This is clearest in the
topic-separated Block~11 views in Figure~\ref{fig:block11_topic_routing_comparison}.
Across all three probes, Block~11 retains highly distributed source usage, partly
concentrated destinations, and a comparable global layout of POS-colored paths.
Active centroids nevertheless vary: slots that are inactive or marginal in one probe
become active sources or destinations in another. We interpret Block~11 as evidence of
topic-conditioned routing modulation within a stable layer-specific memory structure.
Stronger semantic claims require targeted analysis.

\begin{itemize}
	\item \textbf{Block 00}: At the network entrance, the retained probe traces route most source mass through a dominant centroid. Destination routing already fans out over 13 slots, suggesting that early structural sorting begins before source routing has diversified. Source $N_{\mathrm{eff}}$ remains roughly $2.8$--$3.6$, with Gini around $0.96$--$0.97$; Slot~121 captures about $75$--$82$ percent of tokens, while destinations include slots such as S59 for determiners, S107 for nouns, and S29 for punctuation.

	\item \textbf{Block 06}: Mid-depth routing is visibly more distributed, with source and destination bars spread across more than 30 slots. POS-specific paths become clearer: punctuation, determiners, and nouns concentrate in distinct destination slots, and the science probe shows a narrower source distribution than the narrative probe, consistent with repeated technical structure in the input.

	\item \textbf{Block 11}: Among the displayed depths, Block~11 is the high-diversity routing stage. The three topic traces use 60--66 unique source slots, with $N_{\mathrm{eff}}\approx43.5$--$50.0$ and Gini around $0.69$--$0.73$. A small number of self-routes appears, indicating tokens whose source and destination centroids coincide, while the destination side remains moderately concentrated around S62. The politics probe has the broadest source spread, consistent with longer clause structure.

	\item \textbf{Block 15}: The final block compresses most tokens into Slot~109, which receives about $95$--$96$ percent of destination assignments. Source routing remains differentiated, so this is not a collapse of all token content into a single semantic representation; rather, S109 acts as a shared output address reached from topic-specific source paths. The few non-S109 destinations, including S79, S127, and S96, are consistent exceptions across the probes.
\end{itemize}

\begin{table}[htbp]
	\centering
	\caption{Block~15: source slots feeding into the shared destination Slot~109.
		Although the output address is identical across topics, the provenance
		and lexical content differ entirely.}
	\label{tab:block15_s109}
		\begin{tabular}{@{}>{\raggedright\arraybackslash}p{.15\textwidth}>{\raggedright\arraybackslash}p{.30\textwidth}>{\raggedright\arraybackslash}p{.48\textwidth}@{}}
		\toprule
		Topic & Top sources into S109 & Representative tokens carried \\
		\midrule
		Narrative & S23\,(58), S87\,(40), S66\,(17)
		& \texttt{house, forest, arrived, yellow}; \texttt{its, the, nor}; \texttt{She, dog, sentence} \\
		Politics  & S121\,(30), S14\,(19), S116\,(14)
		& \texttt{raised, by, citing, priced}; \texttt{rates, basis, points, Thursday}; \texttt{central, bank, prime} \\
		Science   & S70\,(108), S20\,(37), S100\,(19)
		& \texttt{ode, update, ta}; \texttt{during, with, revealed}; \texttt{lighter, charge, operation} \\
		\bottomrule
		\end{tabular}
	\end{table}

The source-slot distributions summarized in Table~\ref{tab:block15_s109} are
disjoint across topics---narrative draws from 33~unique sources (dominated by
S23/S87), politics from 64 (the most dispersed), and science from 40 (dominated
by S70 alone handling 108~tokens).
The hidden states that accumulate at S109 therefore encode the full upstream
routing history, which is entirely topic-dependent; the LM~head reads a single
address whose content was shaped by 15~blocks of divergent centroid traversals.

\subsubsection{Edge Structure Analysis}

\begin{figure}[htbp]
	\centering
\includegraphics[width=0.9\textwidth]{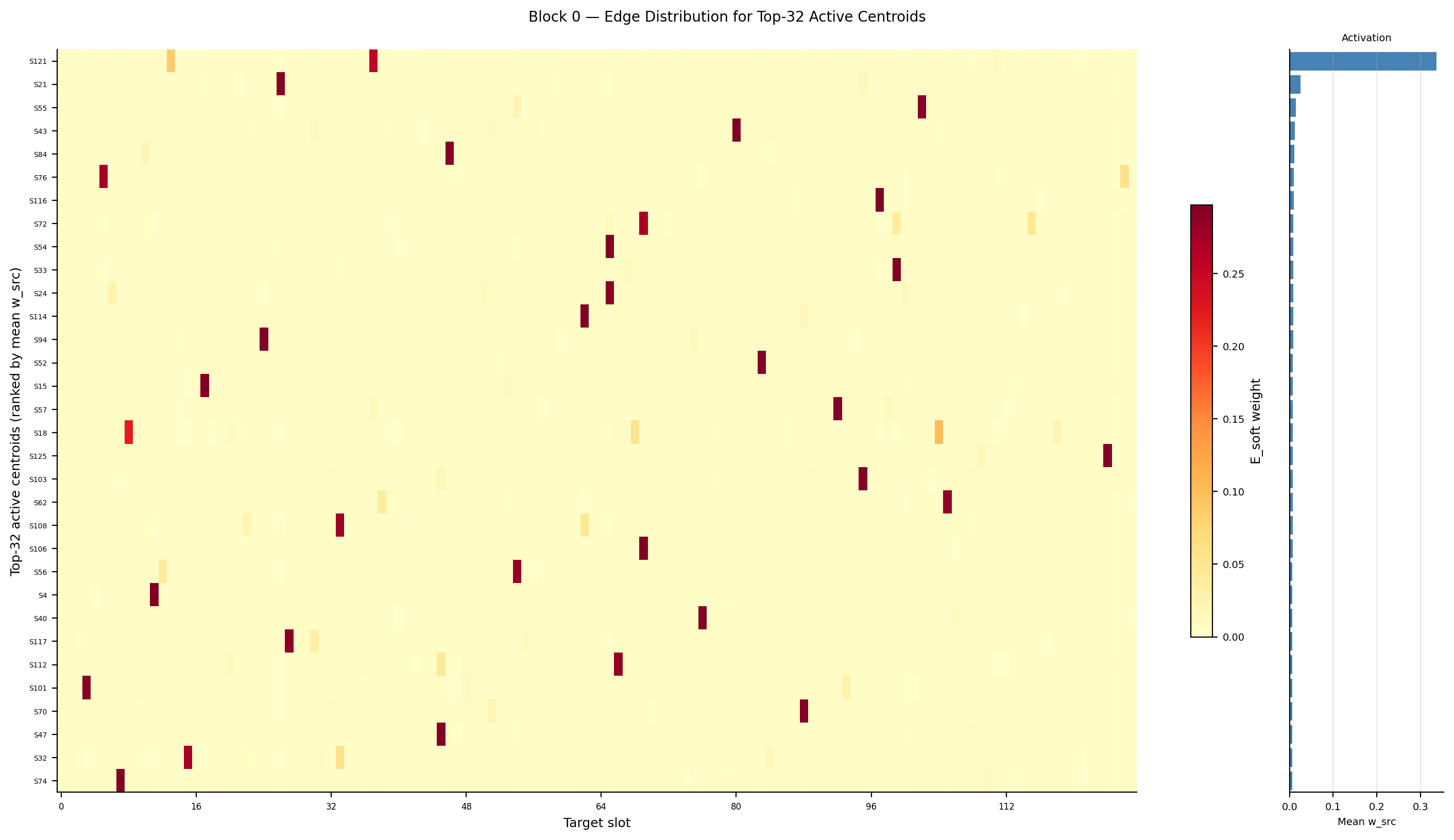}
	\caption{Active edge structure at Block~00. Darker cells indicate stronger routing signal among the top active centroids.}
	\label{fig:edge_active_block00}
\end{figure}

\begin{figure}[htbp]
	\centering
\includegraphics[width=0.9\textwidth]{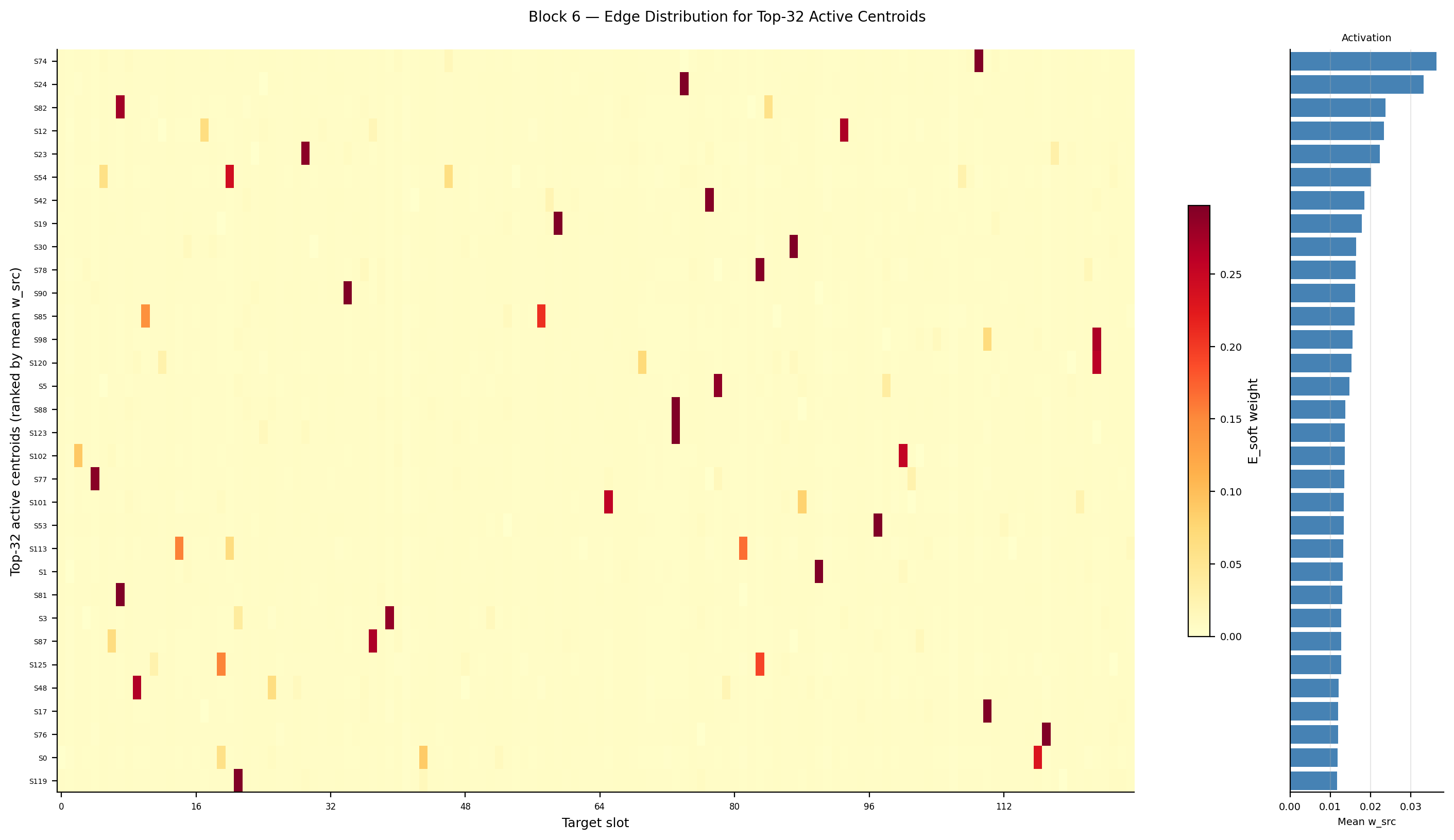}
	\caption{Active edge structure at Block~06.}
	\label{fig:edge_active_block06}
\end{figure}

\begin{figure}[htbp]
	\centering
\includegraphics[width=0.9\textwidth]{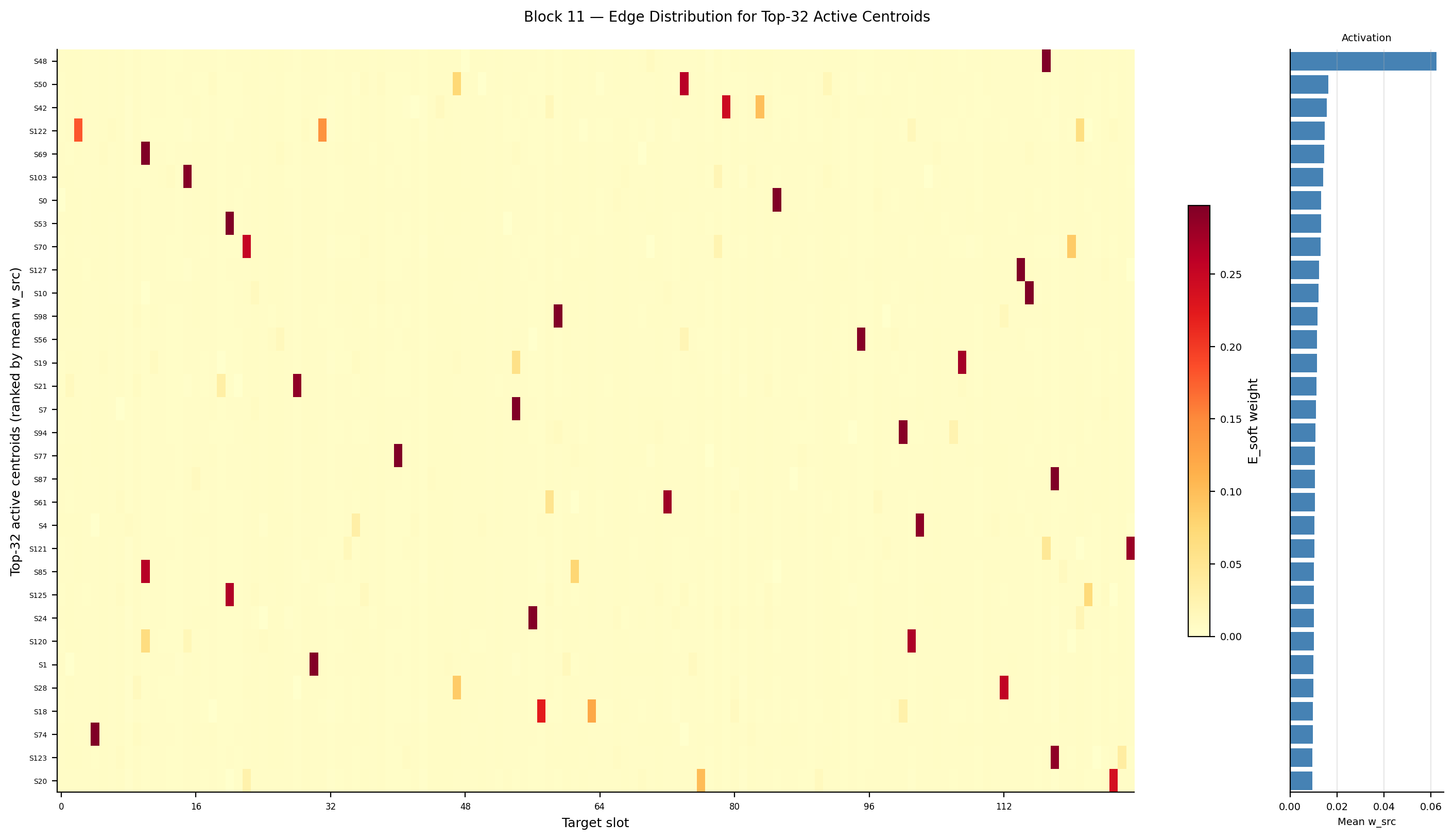}
	\caption{Active edge structure at Block~11.}
	\label{fig:edge_active_block11}
\end{figure}

\begin{figure}[htbp]
	\centering
\includegraphics[width=0.9\textwidth]{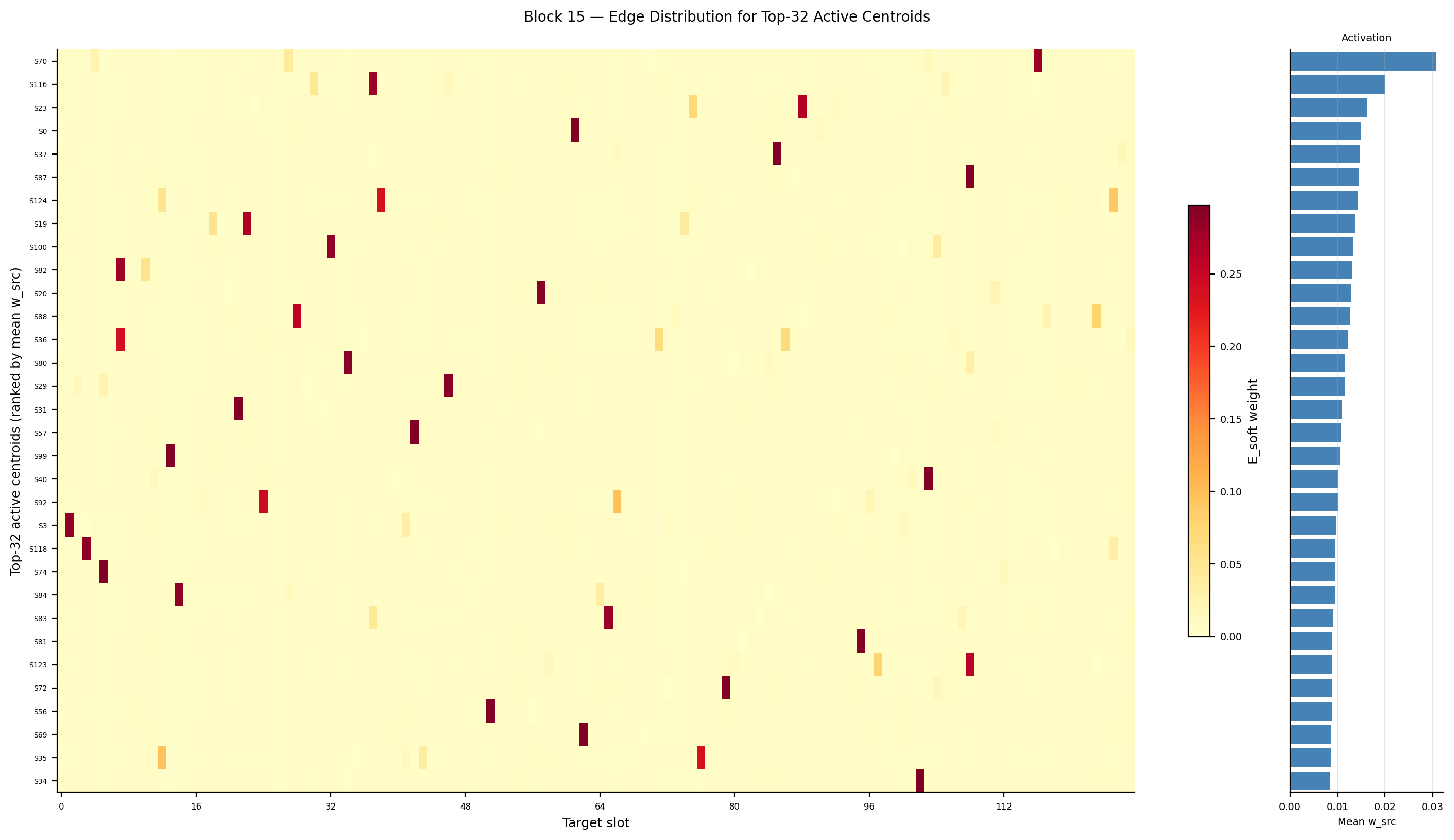}
	\caption{Active edge structure at Block~15. These are the same representative memory blocks used in the routing-flow analysis.}
	\label{fig:edge_active_block15}
\end{figure}

The learned edge matrix complements the routing-flow analysis.
Figures~\ref{fig:edge_active_block00}, \ref{fig:edge_active_block06},
\ref{fig:edge_active_block11}, and~\ref{fig:edge_active_block15} show the
active-edge maps at the same representative depths, restricted to the top 32 active
centroids rather than the full 128-slot graph for readability.

The vertical axis shows the top 32 active centroids/slots, and the horizontal axis
shows target slots. Darker cells indicate stronger edge-routing signals; lighter
cells indicate weaker signals. Most edges tend to route between a particular pair
of slots for a given topic, though the selected edges vary with topic. In some cases,
more than one edge is activated, suggesting that the memory graph can selectively
route parts of a sentence according to the specific context.

\subsection{Sensitivity and Stabilization Analysis}

This section reports retained base v7 diagnostics and selected generation ablations,
alongside development observations that motivated the design. Quantitative statements
draw on preserved artifacts; observations without retained controlled comparisons
serve only as design motivation. The study does not cover every auxiliary objective
or routing hyperparameter, and the corresponding systematic comparisons remain
future work.

\subsubsection{Auxiliary Objectives and Stabilization Diagnostics}

The diagnostics below follow the five memory-specific terms in the training objective
defined in Section~\ref{sec:aux_objectives}:
\begin{equation}
	\begin{aligned}
		\mathcal{L}_{\mathrm{train}}
		&= \mathcal{L}_{\mathrm{LM}}
		+ \lambda_{\mathrm{track}}\mathcal{L}_{\mathrm{track}}
		+ \beta_{\mathrm{ortho}}\mathcal{L}_{\mathrm{ortho}}
		\\
		&\quad
		+ \lambda_{\mathrm{cluster}}\mathcal{L}_{\mathrm{cluster}}
		+ \lambda_{\mathrm{edge}}\mathcal{L}_{\mathrm{edge}}
		+ \lambda_{\mathrm{contrast}}\mathcal{L}_{\mathrm{contrast}}.
	\end{aligned}
\end{equation}
The five terms were introduced to address instabilities observed during development: poor memory
tracking, centroid redundancy, under-utilization, edge collapse, and lack of
transition diversity.

\paragraph{Tracking Loss.}
The tracking loss encourages the source-weighted centroid states to predict the block
state written back into memory:
\begin{equation}
	\mathcal{L}_{\mathrm{track}}
	=
	\left(1 - m\right)
	\operatorname{MSE}
	\!\left(
	\operatorname{sg}\!\left(\mathbf{x}^{(\ell+1)}\right),
	\mathbf{w}_{\mathrm{src}} \widetilde{\mathbf{C}}
	\right).
\end{equation}
Here $m=\sigma(u)$ is the learned write-back momentum. Tracking replaces an earlier
commitment-style formulation inspired by vector-quantized representation learning
\cite{oord2017neural,razavi2019generating,esser2021taming}. Its stop-gradient placement
conflicted with \ac{EMA} write-back, so it is not part of the base v7 objective:
\begin{equation}
	\mathcal{L}_{\mathrm{commit}}
	=
	\operatorname{MSE}\!\left(
	\mathbf{x}^{(\ell+1)},
	\operatorname{sg}\!\left(\mathbf{w}_{\mathrm{src}}\widetilde{\mathbf{C}}\right)
	\right),
\end{equation}
which pulled hidden states toward the memory reconstruction but provided no gradient
to the centroids. Development configurations using it exhibited centroid
collapse, with pairwise cosine similarity rising above $0.6$ within the first epoch.
The tracking loss reverses the stop-gradient placement and couples the objective to
learnable momentum, allowing the model to regulate gradient-based correction relative
to \ac{EMA} drift. Development runs without tracking showed rapid centroid
under-utilization and unstable geometry; the retained artifacts do not support a
controlled comparison.

\paragraph{Orthogonality Loss.}
The orthogonality penalty discourages centroid redundancy by penalizing off-diagonal
entries in the centroid Gram matrix:
\begin{equation}
	\mathcal{L}_{\mathrm{ortho}}
	=
	\frac{1}{F(F-1)}
	\sum_{i\neq j} G_{ij}^{2},
	\qquad
	\mathbf{G}
	=
	\mathbf{C}_{n}\mathbf{C}_{n}^{\top},
	\qquad
	\mathbf{C}_{n,i}
	=
	\frac{\mathbf{C}_{i}}{\lVert\mathbf{C}_{i}\rVert_{2}}.
\end{equation}
Here $G_{ij}$ is the cosine similarity between row-normalized centroids $i$ and $j$.
The loss discourages slots from collapsing onto the same direction while leaving
each centroid free to specialize within the shared hidden-state geometry. In the
retained base v7 run, it is associated with a rapid early reduction in mean pairwise
cosine similarity, from $0.587$ at step $554$ to $0.155$ at step $1664$, followed by a
substantially lower range for the rest of training. This trajectory supports its role
as a stabilizing pressure on centroid geometry; the loss-removal comparison provides
additional evidence.

The no-orthogonality checkpoint e3 has lower lexical diversity on the 23-prompt
suite (Section~\ref{sec:generation_ablations}):
\begin{center}
\small
\begin{tabular}{@{}lrrr@{}}
\toprule
Checkpoint & Distinct-1 $\uparrow$ & Distinct-2 $\uparrow$ & Rep-4 $\downarrow$ \\
\midrule
e0 control & 0.8251 & 0.9568 & 0.0213 \\
e02 control & 0.7665 ($-0.0586$) & 0.9073 ($-0.0495$) & 0.0517 ($+0.0304$) \\
e3 no orthogonality & 0.7369 ($-0.0881$) & 0.8971 ($-0.0598$) & 0.0416 ($+0.0204$) \\
\bottomrule
\end{tabular}
\end{center}
The diversity reduction is consistent with retaining the loss. However, e02, a second
independently trained full-model control, also degrades and has higher repetition
than e3. The comparison provides descriptive support, not an isolated estimate of
the loss's effect across training seeds.

This loss is particularly important because contextual embedding spaces in GPT-2-style
models are generally anisotropic, with high average pairwise cosine similarity in many
layers \cite{ethayarajh2019contextual}. In the memory cell, allowing centroids to
concentrate in a narrow region reduces expressiveness and makes the maintenance
procedure more likely to fall into a trivial configuration.

\paragraph{Clustering Loss.}
The clustering term penalizes an effective source-routed cell count below a target:
\begin{align}
	u_i
	&=
	\frac{
		\frac{1}{BT}\sum_{b,t} w_{\mathrm{src},bti}
	}{
		\sum_j \frac{1}{BT}\sum_{b,t} w_{\mathrm{src},btj} + \epsilon
	},
	\\
	H(\mathbf{u})
	&=
	-\sum_i u_i \log(u_i+\epsilon),
	\qquad
	N_{\mathrm{eff}}
	=
	e^{H(\mathbf{u})},
	\\
	\mathcal{L}_{\mathrm{cluster}}
	&=
	\max\left(
		\frac{N_{\mathrm{target}}}{\max(N_{\mathrm{eff}},1)}
		-1,
		0
	\right).
\end{align}
Here $\mathbf{u}$ is the normalized source-routing usage distribution over the current
batch and sequence positions. With $\lambda_{\mathrm{cluster}}=0.3$ and
$N_{\mathrm{target}}=F/4$, the base v7 mean effective count remains above the target
across logged checkpoints, ranging from $61.6$ to $114.4$ for $F=128$. This one-sided
anti-collapse term guards against under-utilization without forcing an exact cell
count once usage is sufficiently spread out.

\paragraph{Edge Losses.}
Two terms regularize the edge matrix $\mathbf{E}$. The first penalizes masked rows
whose entropy falls below a minimum:
\begin{align}
	P_{ij}
	&=
	\operatorname{softmax}(\mathbf{E}_{i,:}+\mathbf{m}_{i,:})_j,
	\qquad
	m_{ij}
	=
	\begin{cases}
		-\infty, & i=j,\\
		0, & i\neq j,
	\end{cases}
	\\
	H_i
	&=
	-\sum_j P_{ij}\log(P_{ij}+\epsilon),
	\\
	\mathcal{L}_{\mathrm{edge}}
	&=
	\frac{1}{F}
	\sum_{i=1}^{F}
	\max\left(H_{\mathrm{target}}-H_i,0\right).
\end{align}
This keeps each source slot from becoming too sharply restricted to a single target
early in training. In the retained base v7 run, mean edge-row entropy stays close to
the target throughout training, ranging from $3.946$ to $4.352$ for
$H_{\mathrm{target}}=4.0$; the mean largest edge probability at the same checkpoints
remains below $0.283$. These measurements support the term's role as a
minimum-entropy guardrail. Development runs suggested that removing it can produce
low-entropy, near-deterministic off-diagonal routing, but the corresponding ablation
artifacts were not retained.

\paragraph{Contrastive Loss.}
The edge-row contrastive term is
\begin{equation}
	\mathcal{L}_{\mathrm{contrast}}
	=
	\frac{1}{F(F-1)}
	\sum_{i \neq j}
	\langle \mathbf{e}_i,\mathbf{e}_j\rangle,
\end{equation}
where the masked transition distribution for source slot $i$ and its row-wise
$\ell_2$ normalization are
\begin{equation}
	\mathbf{p}_i
	=
	\operatorname{softmax}(\mathbf{E}_{i,:}+\mathbf{m}_{i,:}),
	\qquad
	\mathbf{e}_i
	=
	\frac{\mathbf{p}_i}{\lVert\mathbf{p}_i\rVert_2}
\end{equation}
High entropy within each row does not prevent all rows from
converging to the same distributional shape; the contrastive term addresses this
remaining source of redundancy. In the retained base v7 run, mean edge-row similarity
decreases from $0.397$ at step $554$ to $0.096$ at step $9{,}987$ and $0.094$ at
step $10{,}542$, and reaches
$0.085$ at the final logged checkpoint, step $19{,}974$. This trajectory is consistent
with progressive diversification of outgoing transition patterns under the reported
setting $\lambda_{\mathrm{contrast}}=0.5$.

\paragraph{Loss Weight Configuration.}
The base v7 configuration uses
\begin{equation}
	\begin{aligned}
	\lambda_{\mathrm{track}} &= 1.0,
	&
	\beta_{\mathrm{ortho}} &= 0.05,
	&
	\lambda_{\mathrm{cluster}} &= 0.3,\\
	\lambda_{\mathrm{edge}} &= 0.1,
	&
	\lambda_{\mathrm{contrast}} &= 0.5.
	\end{aligned}
\end{equation}
These fixed coefficients were chosen during development, not through a controlled
loss-weight sweep. Tracking and centroid-geometry regularization were treated as primary
stabilizers; clustering, edge entropy, and edge contrast were used to guard against
under-utilization and transition-graph degeneration. A systematic tracking-loss ablation
and independent variation of all five weights and the clustering threshold
remain future work: controlled sweeps were not retained, leaving marginal effects
and threshold sensitivity unresolved.

\subsubsection{Sensitivity to Memory Size and Routing Hyperparameters}

\paragraph{Number of Memory Slots.}
The choice of $F=128$ slots was informed by development runs, in which larger banks
showed stronger dead-centroid accumulation and less stable utilization. This motivated
a conservative active memory size, but memory size varied alongside other
architectural and training choices: these observations are not a matched sweep or a
scaling result. Determining how memory capacity should scale with model size still
requires a comparison of $F \in \{32,64,128,256\}$ under matched training budgets and
comparable hyperparameter tuning.

\paragraph{Routing Temperature Schedule.}
The soft source-routing weights use a temperature-annealed gravitational distance:
\begin{equation}
	\mathbf{w}_{\mathrm{src}}
	=
	\operatorname{softmax}\!\left(
	\frac{1}{\tau\,d(\mathbf{h}_{n},\mathbf{C}_{n})}
	\right),
	\qquad
	d(\mathbf{h}_{n},\mathbf{c}_i)
	=
	\max\!\left(\varepsilon_{\mathrm{grav}},1-\langle\mathbf{h}_{n},\mathbf{c}_i\rangle\right).
\end{equation}
Temperature is annealed logarithmically from $\tau_{\max}=1.0$ to
$\tau_{\min}=0.1$ over training. The floor $\varepsilon_{\mathrm{grav}}=0.01$ prevents
division by near-zero distances when a hidden state aligns closely with a centroid.
Systematic comparisons with constant or linearly annealed schedules have not been
conducted and remain future work.

\subsubsection{Sensitivity to Initialization and Write-Back Dynamics}

\paragraph{Centroid Initialization.}
Early development configurations initialized centroids using $k$-means over samples
from a pretrained GPT-2 embedding space, without the same orthogonality regularization
and maintenance behavior as base v7. This initialization led to severe packing:
centroids concentrated in a small subregion and stagnated at mean pairwise cosine
similarity above $0.6$, as recorded in the historical
development notes. The packing is consistent with contextual-space
anisotropy \cite{ethayarajh2019contextual}; it made gravitational routing nearly
uniform and left little useful signal for the edge matrix. Spherical initialization,
introduced in a later development configuration and used consistently in base v7,
instead samples centroids as
\begin{equation}
	\mathbf{c}_i
	=
	\frac{\boldsymbol{\epsilon}_i}{\lVert\boldsymbol{\epsilon}_i\rVert_2},
	\qquad
	\boldsymbol{\epsilon}_i \sim \mathcal{N}(\mathbf{0},\mathbf{I}).
\end{equation}
This produces mean cosine similarity near $0.01$ at initialization, allowing contrastive
and orthogonality signals to act from the start of training.
Table~\ref{tab:centroid_maintenance} contrasts the collapsed early memory blocks with
the lower-similarity, more distributed geometry of base v7. This evidence supports
spherical initialization as part of the stabilization recipe, but does not isolate it
from the accompanying losses and maintenance rules: the comparison is development
evidence, not a controlled initialization ablation.

\begin{table}[t]
	\centering
	\small
	\caption{Centroid maintenance and orthogonality observations across development configurations.}
	\label{tab:centroid_maintenance}
	\begin{tabular}{@{}lcccl@{}}
		\toprule
		\textbf{Configuration} & $\beta_{\mathrm{ortho}}$ & \textbf{Init.} & $\cos_{\mathrm{sim}}$ & \textbf{Outcome} \\
		\midrule
		Early k-means A & -- & k-means & $>0.6$ & Collapsed \\
		Early k-means B & -- & k-means & $>0.6$ & Collapsed \\
		\midrule
		Base v7 & 0.05 & sphere & $0.587 \to 0.155$ ($1110$ steps) & Stable \\
		\bottomrule
	\end{tabular}
\end{table}

\paragraph{Gate Initialization.}
The displacement returned by the memory cell is scaled by a learned scalar gate:
\begin{equation}
	\mathbf{x}^{(\ell+1)}
	=
	\mathbf{h}^{(\ell)}
	+
	\sigma(g^{(\ell)})
	\operatorname{LN}^{(\ell)}_{\mathrm{disp}}
	\!\left(
	\mathbf{c}^{(\ell)}_{\mathrm{tgt}}
	-
	\mathbf{c}^{(\ell)}_{\mathrm{src}}
	\right).
\end{equation}
The v7 raw gate parameter is initialized to $g_0=1.0$, giving an effective scale
$\sigma(g_0)\approx0.731$. This intermediate scale keeps the memory path active from
the start of training while limiting the influence of large untrained displacements
on the residual stream. This is implementation rationale; no independent
gate-initialization sweep is reported.

\paragraph{Write-Back Momentum.}
The centroid bank is maintained by an \ac{EMA} write-back:
\begin{equation}
	\mathbf{C}^{\mathrm{new}}_i
	=
	\operatorname{norm}\!\left(
	m\mathbf{C}_i
	+
	(1-m)\overline{\mathbf{r}}_i
	\right),
\end{equation}
where $\overline{\mathbf{r}}_i$ is the mean post-memory block state assigned to slot $i$.
The learnable parameter $u$ sets $m=\sigma(u)$; base v7 initializes it at $u_0=4.6$,
corresponding to $m\approx0.99$. In the training logs, $m$ increases slightly from
approximately $0.990$ at step $554$ to approximately $0.998$ at step $19{,}974$,
remaining close to the high-momentum regime. The gradient path for learning momentum
is described in Appendix~\ref{app:writeback}.

\paragraph{Assignment Strategy.}
Write-back assigns each token to its most strongly weighted source centroid:
\begin{equation}
	i^*_t
	=
	\operatorname*{arg\,max}_{1 \leq j \leq F} w_{\mathrm{src},t,j},
	\qquad
	\overline{\mathbf{r}}_i
	=
	\frac{1}{\max\!\left(|\{t:i^*_t=i\}|,\varepsilon\right)}
	\sum_{t:i^*_t=i}
	\mathbf{r}_t.
\end{equation}
Here $\mathbf{r}_t$ is the post-memory block output, not the pre-memory input used for
source routing. Forward traversal remains soft and differentiable, preserving partial
semantic membership; hard assignments provide clearer centroid targets for decisive
\ac{EMA} updates.


\section{Discussion}

The base v7 results suggest that \ac{GMT} should be understood as a structured
alternative for part of the transformer block update, not simply as a drop-in replacement
for dense feed-forward layers. Its explicit routing and displacement variables make this
transformation more inspectable.

\subsection{Interpretability Implications}

Mechanistic interpretability has shown that transformer computations can often be analyzed through circuits, causal tracing, neuron- or parameter-level localization, sparse feature decompositions, and views of \ac{FFN} layers as memory-like structures \cite{geva2021transformer,meng2022locating,rai2024practical,wang2022interpretability,conmy2023automated,cunningham2023sparse}.
Recent neuron-, layer-, and parameter-activation studies make the post hoc nature of this analysis especially explicit \cite{wang2024activationpatterns,yin2025neuron}.
These methods are valuable, but they usually recover structure after training from dense
activations, interventions, auxiliary probes, or learned decompositions. \ac{GMT} offers
a complementary way to inspect the transformation normally implemented by the \ac{FFN}
branch within a decoder-only transformer: source routing, target routing, centroid usage,
transition weights, gates, and displacement vectors are explicit variables of the forward
computation. The claim is one of structural interpretability, not complete transparency:
several objects otherwise inferred post hoc are available as native architectural quantities.

Tracking memory-cell activation and the displacements produced in each block allows
one to inspect block specialization and how its behavior changes across contexts. The same centroid
region may act as a source in one context and as a destination in another, with
context-dependent meanings and therefore different displacement vectors. Graph traversal
is thus part of the explanation, not only an artifact recovered after the forward pass.

\begin{figure}[htbp]
	\centering
\includegraphics[width=\textwidth]{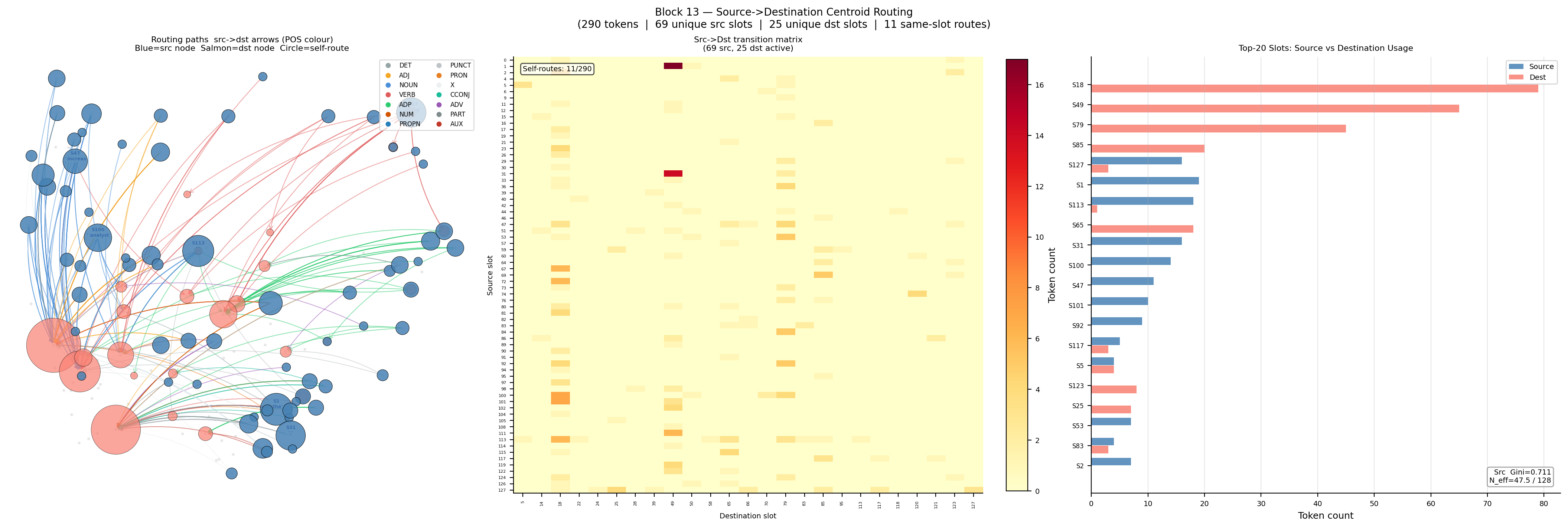}
	\caption{Slot-routing flow in block 13 for a political-text probe, illustrating the source/destination role switching discussed in the interpretability analysis.}
	\label{fig:block13_routing_flow}
\end{figure}

Figure~\ref{fig:block13_routing_flow} illustrates this role switching in block 13,
where Slots 127 and 117 form a two-node pathway with an explicit handoff between distinct
centroid regions.

Slot 127 functions as a predicate source in this probe.
The verbs in the sequence route through it---\textit{citing, signalling, addressing, defending, criticising, arguing, expressing, threatening} and others---with consistently high source entropy $(4.61$--$4.70)$, indicating competitive but stable specialization.
Three of these self-route ($\to$127): \textit{addressed}, \textit{arguing}, \textit{acknowledged}, all verbs that can stand at a clause boundary without a following complement.
The slot is acting as a predicate anchor, holding clause-internal verbs in place while displacing transitional ones outward.

Outward displacement targets Slot 117 selectively: \texttt{' defending'},
\texttt{' criticised'}, and \texttt{' described'}---all transitive verbs that introduce a
prepositional complement---are routed from Slot 127 to Slot 117. Slot 117 then becomes
the source for the relational and directional prepositions that follow those verbs in the
text---\texttt{' in'}, \texttt{' against'}, \texttt{' toward'}, \texttt{' with'},
\texttt{' over'}---displacing them onward to Slot 85. Across this two-node pathway,
the verb centroid handles the predicate class and selectively hands off to the
preposition centroid, which absorbs the displacement and routes onward into the complement
position.

Slot 117's incoming verbs and outgoing prepositions are disjoint POS classes with no
lexical overlap in this probe. The shared centroid origin mediates both transformations,
but their displacement vectors point in opposite directions along distinct trajectories:
incoming tokens from Slot 127 follow one directional axis, while the structurally different
outgoing token class follows another.

\subsection{Limitations}

The present evidence should be interpreted as an architectural proof of concept rather
than as a comprehensive performance study. The dense GPT-style baseline has 20.8M more
parameters than the base v7 \ac{GMT} model, so the comparison only imperfectly isolates
the feasibility of the \ac{FFN}-to-memory substitution. A parameter-matched baseline,
multiple random seeds, and broader controlled ablations of routing, write-back,
auxiliary losses, and centroid maintenance beyond the selected diagnostics are needed
before stronger conclusions can be drawn about efficiency or robustness. The benchmark
set is also limited: the language-model results show stable training and close downstream
behavior in the evaluated setting, but
they do not establish broad task generality, scaling behavior, or superiority over dense
transformer baselines.

\section{Conclusion and Future Work}
\label{sec:conclusion}

\subsection{Conclusion}

This paper introduced the base v7 \ac{GMT}, a decoder-only transformer that replaces
the feed-forward branch with a graph-structured memory cell while preserving the
surrounding transformer architecture. Its post-attention transformation consists of
source routing, one-hop graph traversal, token-conditioned target selection, and a
gated displacement returned to the residual stream. The architecture exposes routing,
centroid usage, transition structure, displacement vectors, gate values, and
memory-maintenance dynamics directly in the forward computation, providing concrete
objects for analysis rather than structure inferred only after training.

The model trains stably and preserves meaningful downstream behavior with a smaller
parameter budget than the dense GPT-style baseline, although it trails that baseline
in validation loss and perplexity. The main result is an architectural proof of concept
for viable, structurally interpretable graph-mediated token transformation, not
superiority over dense transformers. Parameter-matched baselines, multi-seed studies,
and mechanistic interventions remain necessary next steps.

\subsection{Future Work}

Only the base v7 architecture is evaluated here. The following directions address
questions left open by that evaluation and are hypotheses for subsequent studies,
not additional results.

\subsubsection{Evaluation and Mechanistic Validation}

Before extending the architecture, the current mechanism needs stronger evidence from
parameter-matched dense baselines, repeated runs over multiple random seeds, and
interventions on the learned graph and displacement path. At evaluation time, the learned
edge matrix could be compared against
identity, uniform, shuffled, or randomized alternatives, while the displacement
contribution could be scaled over a range of fixed $\alpha$ values as
\begin{equation}
	\begin{aligned}
	\mathbf{x}^{(\ell+1)}_t(\alpha)
	&=
	\mathbf{h}^{(\ell)}_t
	+
	\alpha\,\mathbf{d}^{(\ell)}_t,
	\\
	\mathbf{d}^{(\ell)}_t
	&=
	\sigma\!\left(g^{(\ell)}\right)
	\operatorname{LN}^{(\ell)}_{\mathrm{disp}}
	\!\left(
	\mathbf{c}^{(\ell)}_{\mathrm{tgt},\,t}
	-
	\mathbf{c}^{(\ell)}_{\mathrm{src},\,t}
	\right),
	\qquad \alpha \geq 0.
	\end{aligned}
\end{equation}
The base v7 update corresponds to $\alpha=1$. These controls would help distinguish
whether the observed behavior comes from the graph structure itself, from the centroid bank acting
as a generic memory, or simply from residual capacity elsewhere in the model.

\subsubsection{Hybrid Dense and Graph-Memory Computation}

The base v7 model deliberately replaces the dense \ac{FFN} in every transformer block.
A hybrid could instead augment dense token-wise computation with graph memory,
retaining the semantic displacement path and adding a controlled dense path:
\begin{equation}
	\begin{aligned}
	\mathbf{z}
	&=
	\mathbf{x}
	+
	\alpha_{\mathrm{mem}}\mathbf{d}_{\mathrm{mem}},
	\\
	\mathbf{y}
	&=
	\mathbf{z}
	+
	\alpha_{\mathrm{ffn}}
	\operatorname{FFN}\!\left(\operatorname{LN}(\mathbf{z})\right),
	\end{aligned}
\end{equation}
where $\mathbf{x}$ denotes the post-attention state entering the second sublayer,
$\mathbf{z}$ is the memory-updated intermediate state, and the residual scales may
be learned per block or per channel. The setting $\alpha_{\mathrm{mem}}=1$ and
$\alpha_{\mathrm{ffn}}=0$ recovers the memory-only second sublayer studied in the
base v7 model, while nonzero $\alpha_{\mathrm{ffn}}$ defines a controlled hybrid.
This would test whether graph memory is most useful as a structured transition
operating alongside ordinary nonlinear channel mixing, rather than as the only
source of token-wise transformation.

\subsubsection{Multi-Hop Graph Traversal}

The base v7 model uses a deliberately minimal single step from source routing through
the edge matrix to target routing. A multi-hop extension would test whether deeper
concept composition can be supported by propagating the routing distribution over
multiple graph steps:
\begin{equation}
	\begin{aligned}
	\mathbf{w}^{(0)}_t
	&=
	\mathbf{w}_{\mathrm{src},\,t},
	\\
	\mathbf{w}^{(k)}_t
	&=
	\operatorname{softmax}
	\!\left(
	\mathbf{w}^{(k-1)}_t\mathbf{A}
	+
	\mathbf{s}_{\mathrm{ctx}}(\mathbf{h}_t)
	\right),
	\qquad k=1,\ldots,K,
	\\
	\mathbf{d}_{K,t}
	&=
	\sigma\!\left(g\right)\,
	\operatorname{LN}_{\mathrm{disp}}
	\!\left(
	(\mathbf{w}^{(K)}_t-\mathbf{w}^{(0)}_t)\widetilde{\mathbf{C}}
	\right).
	\end{aligned}
\end{equation}
Here $\mathbf{A}$ denotes the row-normalized transition graph and
$\mathbf{s}_{\mathrm{ctx}}(\mathbf{h}_t)$ denotes the token-conditioned context score.
For $K=1$, this recovers the base v7 one-hop target routing and gated
target-minus-source displacement readout; values $K>1$ define a future extension.
Repeated graph propagation may increase compositional depth, but it can also over-smooth
the routing distribution or weaken the signal carried by later hops. Hop-wise residual
gates, entropy diagnostics, and explicit
comparisons between final-hop and multi-scale readout would therefore be needed.

\subsubsection{Memory Maintenance and Write-Back}

The base v7 model's hard source assignments for write-back preserve discrete slot
identity but can be brittle when source routing is uncertain. They should be compared
with soft, top-$k$, and confidence-gated write-back rules. If
$I_k(t)$ denotes the top-$k$ source slots for token $t$, one possible update family is
\begin{equation}
	\begin{aligned}
	a^{(k)}_{t,i}
	&=
	\begin{cases}
	\dfrac{w_{\mathrm{src},t,i}}
	{\sum_{j\in I_k(t)} w_{\mathrm{src},t,j}},
	& i\in I_k(t), \\
	0,
	& \text{otherwise},
	\end{cases}
	\\
	\bar{\mathbf{r}}^{(k)}_i
	&=
	\frac{
	\sum_t \gamma_t a^{(k)}_{t,i}\mathbf{r}_t
	}{
	\max\!\left(
	\sum_t \gamma_t a^{(k)}_{t,i},
	\varepsilon
	\right)
	}.
	\end{aligned}
\end{equation}
Here $\gamma_t \in [0,1]$ is an optional confidence term derived from routing entropy
or top-two margin. Setting $k=1$ and $\gamma_t=1$ recovers the base v7 hard write-back
target before applying the same \ac{EMA} and centroid-renormalization update described
above. Separating train-time maintenance from evaluation-time and generation-time
behavior will also be important, since online mutation can otherwise make it difficult
to isolate the effect of the learned parameters from the effect of state updates.

\subsubsection{Controlled Memory Flexibility}

Controlled extensions would test whether the centroid graph can gain expressive capacity
while preserving the base model's displacement-based interpretability. Semantic slot
identity is central to the \ac{GMT} interpretation, so any separation between routing geometry and readout
content should be constrained to avoid turning the graph into a generic key-value lookup
table. Examples include a residual value component initialized at zero, a shared semantic
slot embedding from which routing and readout views are derived, multi-head memory spaces, or
token-conditioned low-rank edge modulation.

\subsubsection{Vision and Anomaly Detection}

For vision models, particularly in anomaly detection, memory-bank methods have proved
effective in industrial inspection by storing normal patch-level features extracted from
pretrained visual backbones and scoring test patches
by their distance from nominal feature support
\cite{gong2019memorizing,defard2021padim,roth2022totalrecall}. A vision-oriented
\ac{GMT} variant could replace a flat memory bank with a compact graph of learned
centroids trained on normal patch embeddings. In such a setting, anomaly scores could be
derived not only from distance to the nearest normal centroid, but also from routing
entropy, unlikely centroid transitions, and unusually large source-to-target
displacements. This application remains speculative: the v7 base model is
evaluated only as a language model, and any claim about visual anomaly detection would
require a dedicated vision architecture and evaluation on industrial anomaly benchmarks
such as MVTec AD \cite{bergmann2019mvtec}.

\section*{Statements and Declarations}

\paragraph{Funding}
This research was self-funded by the authors. No external funding was received for
conducting the study.

\paragraph{Competing interests}
The authors declare no competing interests relevant to this work.

\paragraph{Author contributions}
Nicola Zanarini conceived the method, developed the initial training and experimental
code, provided computational infrastructure, supervised the research, prepared the
first manuscript draft, and reviewed the manuscript. Niccol\`o Ferrari refactored
the research code, developed the public implementation, provided computational
infrastructure, and wrote the current manuscript. Evelina Lamma provided academic
supervision and reviewed the manuscript.

\paragraph{Data availability}
The study used the publicly available training and evaluation datasets identified
in the Experimental Setup, including OpenWebText, ARC, HellaSwag, PIQA, and WinoGrande.

\paragraph{Code and model availability}
The public implementation is available at
\url{https://github.com/Nemesis533/GMT-GraphMemoryTransformer}.
The base GMT v7 checkpoint is available at
\url{https://huggingface.co/NicolaZanarini/gmt-v7-base}.

\clearpage
\bibliography{biblio}

\clearpage
\appendix

\section{Reference V7 Realization Details}

This appendix specifies the reference base-v7 write-back, tracking, and maintenance behavior needed to interpret the method and results.

\subsection{Online Write-Back Realization}
\label{app:writeback}

In the reference base-v7 realization, write-back (Section~\ref{sec:writeback}) runs under \texttt{torch.no\_grad()} on every forward pass, not only after optimizer steps. It receives the source-routing weights and post-block state, which already includes the memory-cell displacement. The current scalar value of $m$ is used for this online state update without backpropagating through centroid assignment. Momentum is learned through the tracking objective (Section~\ref{sec:tracking_loss}), where $(1-m)$ modulates the pressure for source-weighted centroids to predict the written state.

\subsection{Tracking Target and Hard Assignment}

The v7 tracking loss and write-back use the same post-block state. The argmax source assignments and per-centroid means defined in Section~\ref{sec:writeback} are realized with one-hot masks before the EMA update.

\subsection{Periodic Maintenance Policy}

The optimizer-step schedule specified in Section~\ref{sec:training_maintenance} triggers centroid maintenance every
\begin{equation}
K_{\mathrm{maint}} = 110
\end{equation}
global steps. At each event, dead-centroid reset precedes similarity-based merging.

\subsection{Reset and Merge Realization}

Reset uses the current-batch or random-vector replacements specified in Section~\ref{sec:centroid_maintenance}. Similarity-based merging repurposes the less-used centroid of an overly similar mature pair without averaging it with its partner. Its replacement is sampled from normalized token states in the current batch.

\end{document}